\documentclass[10pt]{article} 
\usepackage[preprint]{tmlr}
\usepackage{pifont}
\usepackage{enumitem}

\usepackage{amsmath,amsfonts,bm}

\def\eqref#1{equation~\ref{#1}}

\def\1{\bm{1}}

\def\ra{{\textnormal{a}}}

\def\rx{{\textnormal{x}}}

\def\rva{{\mathbf{a}}}

\def\erva{{\textnormal{a}}}

\def\ervx{{\textnormal{x}}}

\def\rmA{{\mathbf{A}}}

\def\vmu{{\bm{\mu}}}
\def\vtheta{{\bm{\theta}}}
\def\va{{\bm{a}}}

\def\ve{{\bm{e}}}

\def\vx{{\bm{x}}}
\def\vy{{\bm{y}}}
\def\vz{{\bm{z}}}

\def\eva{{a}}

\def\mA{{\bm{A}}}
\def\mB{{\bm{B}}}

\def\mH{{\bm{H}}}
\def\mI{{\bm{I}}}
\def\mJ{{\bm{J}}}

\def\mU{{\bm{U}}}

\def\mW{{\bm{W}}}
\def\mX{{\bm{X}}}

\def\mSigma{{\bm{\Sigma}}}

\DeclareMathAlphabet{\mathsfit}{\encodingdefault}{\sfdefault}{m}{sl}
\SetMathAlphabet{\mathsfit}{bold}{\encodingdefault}{\sfdefault}{bx}{n}
\newcommand{\tens}[1]{\bm{\mathsfit{#1}}}
\def\tA{{\tens{A}}}

\def\tX{{\tens{X}}}

\def\gG{{\mathcal{G}}}

\def\sA{{\mathbb{A}}}
\def\sB{{\mathbb{B}}}

\def\sR{{\mathbb{R}}}
\def\sS{{\mathbb{S}}}

\def\emA{{A}}

\newcommand{\etens}[1]{\mathsfit{#1}}

\def\etA{{\etens{A}}}

\newcommand{\E}{\mathbb{E}}

\newcommand{\R}{\mathbb{R}}

\newcommand{\KL}{D_{\mathrm{KL}}}
\newcommand{\Var}{\mathrm{Var}}

\newcommand{\Cov}{\mathrm{Cov}}

\newcommand{\normltwo}{L^2}
\newcommand{\normlp}{L^p}

\newcommand{\parents}{Pa} 

\usepackage{hyperref}
\usepackage{url}
\usepackage[textsize=scriptsize]{todonotes}
\setuptodonotes{fancyline, color=yellow!20}
\newcommand*\samethanks[1][\value{footnote}]{\hypersetup{linkcolor=blue}\footnotemark[#1]}

\title{{\em How to think step-by-step:} A mechanistic understanding of chain-of-thought reasoning}

\author{\name Subhabrata Dutta\thanks{Equal contribution.} \email subha0009@gmail.com \\
      \addr IIT Delhi, India
      \AND
      \name Joykirat Singh\samethanks \email joykiratsingh18@gmail.com \\
      \addr Independent\thanks{Work done as Research Assistant at IIT Bombay}
      \AND
      \name Soumen Chakrabarti \email soumen@cse.iitb.ac.in\\
      \addr IIT Bombay, India
      \AND
      \name Tanmoy Chakraborty \email tanchak@ee.iitd.ac.in \\
      \addr IIT Delhi, India}

\def\month{MM}  
\def\year{YYYY} 
\def\openreview{\url{https://openreview.net/forum?id=XXXX}} 

\begin{document}

\maketitle

\begin{abstract}
Despite superior reasoning prowess demonstrated by Large Language Models (LLMs) with Chain-of-Thought (CoT) prompting, a lack of understanding prevails around the internal mechanisms of the models that facilitate CoT generation. This work investigates the neural sub-structures within LLMs that manifest CoT reasoning from a mechanistic point of view. From an analysis of {\color{black} Llama-2} 7B applied to multistep reasoning over fictional ontologies, we demonstrate that LLMs deploy multiple parallel pathways of answer generation for step-by-step reasoning. These parallel pathways provide sequential answers from the input question context as well as the generated CoT. {\color{black}We observe a functional rift in the middle layers of the LLM. Token representations in the initial half remain strongly biased towards the pretraining prior, with the in-context prior taking over in the later half. This internal phase shift manifests in different functional components: attention heads that write the answer token appear in the later half, attention heads that move information along ontological relationships appear in the initial half, and so on.} To the best of our knowledge, this is the first attempt towards mechanistic investigation of CoT reasoning in LLMs.
\end{abstract} 

\section{Introduction}
\label{sec:intro}

{\color{black} Recent advancements with Large Language Models (LLMs) demonstrate their remarkable reasoning prowess with natural language across a diverse set of problems~\citep{openai2024gpt4, AutoCoT, chen2023pot}.  Yet, existing literature fails to describe the neural mechanism within the model that implements those abilities -- how they emerge from the training dynamics and why they are often brittle against even unrelated changes.  One of these capabilities of LLMs that has boosted their potential in complex reasoning is {\em Chain-of-Thought} (CoT) prompting ~\citep{CoT-prompting-2022, AutoCoT}. Instead of providing a direct answer to the question, in CoT prompting, we expect the model to generate a verbose response, adopting a step-by-step reasoning process to reach the answer. 
Despite the success of eliciting intermediate computation and their diverse, structured demonstration, the exact mechanism of CoT prompting remains mysterious.  Prior attempts have been made to restrict the problem within structured, synthetic reasoning to observe the LLM's CoT generation behavior~\citep{saparov2023prontoqa}; context perturbation towards causal modeling has also been used~\citep{tan2023causal-CoT}. A few recent approaches seek to associate the emergence of CoT reasoning ability in LLMs with localized structures in the pertaining data~\citep{prystawski2024think, wang2023reasoning}. However, these endeavors produce only an indirect observation of the mechanism; the underlying neural algorithm implemented by the LLM remains in the dark.}

Recent developments in the mechanistic interpretation of Transformer-based models provide hope for uncovering the neural `algorithms' at work inside LLMs~\citep{elhage2021mathematical, nanda2022comprehensive}. Typically, mechanistic methods seek to build a causal description of the model; starting from the output, they localize important components of the model (e.g., an attention head or a multilayer perceptron (MLP) block) by activation patching~\citep{wang2023interpretability, zhang2023patching}. However, there are some implicit challenges in reverse-engineering CoT-prompting in foundational models. The capability of generating CoT is largely dependent on model scale~\citep{wei2022emergent, saparov2023prontoqa}. On the other hand, reverse-engineering a large model becomes a wild goose chase due to the {\em hydra effect}~\citep{mcgrath2023hydra} --- neural algorithmic components within LLMs are adaptive: once we `switch off' one functional component, others will pitch in to supply the missing functionality. Furthermore, in most real-world problems involving multi-step reasoning, there are implicit knowledge requirements. LLMs tend to memorize factual associations from pretraining as key-value caches using the MLP blocks~\cite{geva-etal-2021-transformer, rome}. Due to the large number of parameters in MLP blocks and their implicit polysemanticity, interpretation becomes extremely challenging. 

In this paper, we seek to address these challenges, thus shedding light on the internal mechanism of Transformer-based LLMs while they perform CoT-based reasoning. Going beyond the typical toy model regime of mechanistic interpretation, we work with {\color{black} Llama-2} 7B~\citep{touvron2023llama}. To minimize the effects of MLP blocks and focus primarily on reasoning from the provided context, we make use of the PrOntoQA dataset~\citep{saparov2023prontoqa} that employs ontology-based question answering using fictional entities (see Figure~\ref{fig:full-circuit-diagram} for an example). Specifically, we dissect CoT-based reasoning on fictional reasoning as a composition of a fixed number of subtasks that require decision-making, copying, and inductive reasoning (\ding{42} Section~\ref{sec:task-composition}).

We draw on three prominent techniques for investigating neural algorithms, namely, activation patching~\citep{nanda2022comprehensive}, probing classifiers~\citep{2022-probing}, and logit lens~\citep{logit-lens} to disentangle different aspects of the neural algorithm implemented by {\color{black} Llama-2} 7B. We find that despite the difference in reasoning requirement of different subtasks, the sets of attention heads that implement their respective algorithms enjoy significant intersection (\ding{42} Section~\ref{subsec:task-specific-heads}). Moreover, they point towards the existence of mechanisms similar to induction heads~\citep{olsson2022icl-induction-heads} working together. In the initial layers of the model (typically, from first to 16th decoder blocks in {\color{black} Llama-2} 7B), attention heads conduct information transfer between ontologically related fictional entities; e.g., for the input {\em numpuses are rompuses}, this mechanism copies information from {\em numpus} to {\em rompus} so that any pattern involving the former can be induced with the latter (\ding{42} Section~\ref{sec:token-mixing}). Interestingly, in this same segment of the model, we find a gradual transition in the residual stream representation from pretraining before in-context prior, i.e., the contextual information replaces the bigram associations gathered via pretraining (\ding{42} Section~\ref{subsec:in-context}).

Following this, we seek to identify the pathways of information that are responsible for processing the answer and write it to the output residual stream for each subtask. To deal with the abundance of backup circuits~\citep{wang2023interpretability}, we look for such parallel pathways simultaneously without switching off any of them. We find that multiple attention heads simultaneously write the answer token into the output, though all of them appear at or after the 16th decoder block (\ding{42} Section~\ref{subsec:answer-writer}). These answer tokens are collected from multiple sources as well (i.e., from the few-shot context, input question, and the generated CoT context), pointing towards the coexistence of multiple neural algorithms working in parallel.

Our findings supply empirical answers to a pertinent open question about whether LLMs actually rely on CoT to answer questions~\citep{tan2023causal-CoT, lampinen-etal-2022-CoT-no-causal}: the usage of generated CoT varies across subtasks, and there exists parallel pathways of answer collection from CoT as well as directly from the question context (\ding{42}~Section~\ref{subsec:answer-collection}).  Here again, we observe the peculiar functional rift within the middle of the model: answer tokens, that are present in the few-shot examples but contextually different from the same token in the question, 
are used as sources by attention heads, primarily before the 16th decoder block. To the best of our knowledge, this is the first-ever in-depth analysis of CoT-mediated reasoning in LLMs in terms of the neural functional components. Code and data~\footnote{\url{https://github.com/joykirat18/How-To-Think-Step-by-Step}} are made available publicly.

\section{Related work}
\label{sec:related-work}

In this section, we provide an overview of literature related to this work, primarily around different forms of CoT reasoning with Transformer-based LMs, theoretical and empirical investigations into characteristics of CoT, and mechanistic interpretability literature around language models.

\citet{nye2021scratchpad} first showed that, instead of asking to answer directly, letting an autoregressive Transformer generate intermediate computation, which they called {\em scratchpads}, elicits superior reasoning performance. They intuitively explained such behavior based on the observation that, in a constant depth-width model with \(\mathcal{O}(n^2)\) complexity attention, it is not possible to emulate an algorithm that requires super-polynomial computation; by writing the intermediate answer, this bottleneck is bypassed. \citet{CoT-prompting-2022} demonstrated that LLMs enabled with unstructured natural language expressions of intermediate computations, aka CoT, can demonstrate versatile reasoning capabilities.

Multiple recent attempts have been made toward a deeper understanding of CoT, both empirically and theoretically. \citet{feng2023CoT-with-circuits} employed circuit complexity theory to prove that --- (i)~Transformers cannot solve arithmetic problems with direct answers unless the model depth grows super-polynomially with respect to problem input size, and (ii)~Constant depth Transformers with CoT generation are able to overcome the said challenge. \citet{liu2022shortcut-automata} showed that Transformers can learn {\em shortcut} solutions to simulate automata with standard training (without CoT); however, they are statistically brittle, and recency-biased scratchpad training similar to that of \citet{nye2021scratchpad} can help with better generalizability and robustness.
\citet{saparov2023prontoqa} dissected the behavior of LLMs across scale on synthetically generated multistep reasoning tasks on true, false, and fictional ontologies. In a somewhat counterintuitive finding, \citet{lampinen-etal-2022-CoT-no-causal} observed that even if an LLM is prompted to generate an answer followed by an explanation, there is a significant improvement over generating an answer without any explanation. While this might point toward a non-causal dependence between the explanation and the answer, \citet{tan2023causal-CoT} showed that LLMs do utilize, at least partially, the information generated within the intermediate steps such that the answer is causally dependent on the CoT steps.
{\color{black} \citet{prystawski2024think} theoretically associated the emergence of CoT with localized statistical structures within the pertaining data. \citet{wang2023reasoning} investigated reasoning over knowledge graph to characterize the emergence of CoT. They identified that reasoning capabilities are primarily acquired in the pertaining stage and not downstream fine-tuning while empirically validating \citet{prystawski2024think}'s theoretical framework in a more grounded manner. \citet{bi2024program} found that while reasoning via program generation, LLMs tend to favor an optimal complexity of the programs for best performance.}

The above-mentioned studies either 1)~dissect the model behavior under controllable perturbations in the input problem or 2)~construct theoretical frameworks under different assumptions to prove the existence of certain abilities/disabilities. Mechanistic interpretability techniques go one step deeper and seek to uncover the neural algorithm deployed by the model to perform a certain task. \citet{elhage2021mathematical} studied one-, two-, and three-layer deep, attention-only autoregressive Transformers and their training dynamics. A crucial finding of their investigation was the existence of {\em induction heads} --- a composition of two attention heads at different layers that can perform pattern copying from context. \citet{olsson2022icl-induction-heads} empirically observed the simultaneous emergence of induction heads and in-context learning in the training dynamics of similar toy Transformer models. Similar analyses on phenomena like polysemanticity, superposition, and memorization have been performed in the toy model regime~\citep{elhage2022toy-superposition,henighan2023superposition}. Beyond these toy models, \citet{wang2023interpretability} analyzed circuits in GPT-2 responsible for Indirect Object Identification. \citet{wu2023interpretability} proposed a causal abstraction model to explain price-tagging in the Alpaca-7B model: given a range of prices and a candidate price, the task is to classify if the candidate price falls within the given range. Extrapolating mechanistic observations from the regime of toy models to `production' LLMs is particularly challenging; with billions of parameters and deep stacking of attention heads, identifying head compositions that instantiate an algorithm is extremely difficult. Furthermore, as \citet{mcgrath2023hydra} suggest, different components of the LLM are loosely coupled and adaptive in nature, for which they coined the term {\em hydra effect} --- ablating an attention layer may get functionally compensated by another. \citet{wang2023interpretability} also identified similar information pathways called {\em backup circuits} that can take over once the primary circuits are corrupted.

\section{Background}
\label{sec:background}

In this section, for completeness, we briefly introduce the concepts and assumptions necessary for delving into the interpretation of CoT reasoning.

The {\bf Transformer architecture}~\citep{transformer} consists of multiple, alternated attention and MLP blocks, preceded and followed by an embedding projection and a logit projection, respectively. Given the vocabulary as \(V\), we denote the embedded representation of the \(i\)-th token \(s_i\in V\) in an input token sequence \(S\) of length \(N\) (i.e., the sum of embedding projections and position embeddings in case of additive position encoding) as \(\vx^i_0 \in \sR^{d}\), where \(d\) is the model dimension. 
We denote the content of the residual stream corresponding to the \(i\)-th token, input to the \(j\)-th decoder block, as \(\vx^i_{j-1}\), which becomes \(\tilde{\vx}^i_{j-1}\) after the attention layer reads and writes on the residual stream. The initial content of the residual stream is the same as the token embedding \(\vx_0^i\). Assuming \(H\) number of heads for each attention layer, the operation of \(k\)-th attention head at \(j\)-th decoder block on the \(i\)-th token's residual stream is denoted as \(\vy^i_{j,k}=h_{j,k}(\vx^i_{j-1})\), \(\vy^i_{j,k}\in \sR^{d}\). 
Then \(\tilde{\vx}^i_{j}=\vx^i_{j-1}+\sum_{k}\vy^i_{j,k}\) is the content of the residual stream immediately after the attention operations. Each attention head \(h_{j,k}\) is parameterized using four projection matrices: query, key and value projection matrices, denoted by \(\mW^{j,k}_Q, \mW^{j,k}_K, \mW^{j,k}_V \in \sR^{d\times \frac{d}{H}}\), respectively, and output projection matrix \(\mW^{j,k}_O\in \sR^{\frac{d}{H} \times d}\). Note that in the case of {\color{black} Llama} models, the position encoding is incorporated via rotation of query and key projections before computing dot-product~\citep{su2023roformer}; we omit this step for brevity. Similarly, the action of the MLP block can be expressed as \(\vz^i_{j}=\operatorname{MLP}_{j}(\tilde{\vx}^i_{j})\), with \(\vx^i_{j}=\tilde{\vx}^i_{j}+\vz^i_{j}\) denoting the content of the residual stream after decoder block \(j\). After processing the information through \(L\) number of decoder blocks, a feedforward transformation \(\mU\in \mathbb{R}^{d\times |V|}\), commonly known as the {\em unembedding} projection, maps the content of the residual stream into the logit space (i.e., a distribution over the token space). Given a sequence of input tokens \(S=\{s_1, \cdots, s_i\}\), the autoregressive Transformer model predicts a new output token \(s_{i+1}\):
\begin{align}
    s_{i+1} = \arg \max_{\vx^{i}_\text{logit}} \operatorname{LM}(\vx^{i}_\text{logit}|S)
\end{align}

{\bf Fictional ontology}-based reasoning, as proposed by~\citet{saparov2023prontoqa} as PrOntoQA, provides a tractable approach to dissect CoT generation. The reasoning problem is framed as question-answering on a tree-based ontology of fictional entities (see Appendix~\ref{app:subsec:prontoqa-example} for examples of ontologies and reasoning problems framed). This eases two major challenges in our case: (i)~Mechanistic interpretation requires input or activation perturbation and recording the results of such perturbations. {\color{black}With CoT, one then needs to repeat such perturbation process for all the constituent subtasks.} Unlike free-form CoT reasoning, PrOntoQA provides a clearly demarcated sequence of successive steps that can be analyzed independently. (ii)~The solution to most real-world reasoning problems heavily requires factual knowledge, so much so that a sound reasoning process might get misled by incorrect fact retrieval. {\color{black}A fictional ontology ensures zero interference between the entity relationships presented in the question and the world-knowledge acquired by the model in the pertaining stage. This further minimizes the involvement of the MLP layers in the neural mechanisms implemented by the model as they typically serves as the parametric fact memory of an LM~\citep{geva-etal-2021-transformer, rome}.}  {\color{black} Additionally, PrOntoQA provides reasoning formulation over false ontologies (see example in Appendix~\ref{app:subsec:prontoqa-example}). False ontology grounds the reasoning over statements that are false in the real world. An important demarcation between fictional and false ontological reasoning is that while the former minimizes the effects of factual associations memorized as pretraining prior, the latter requires the LM to actively eclipse such memorized knowledge to solve the problem successfully.}

{\bf Circuits}, in mechanistic interpretability research, provide the abstractions of interpretable algorithms implemented by the model within itself. Typically, a \emph{circuit} is a subgraph of the complete computational graph of the model, responsible for a specific set of tasks. We primarily follow the notation adopted by~\citet{wang2023interpretability}, with nodes defined by model components like attention heads and projections and edges defined by interactions between such components in terms of attention, residual streams, etc.

{\bf Activation patching} is a common method in interpretability research. Activation patching begins with two forward passes of the model, one with the actual input and another with a selectively corrupted one. The choice of input corruption depends on the task and the type of interpretation required. For example, consider the following Indirect Object Identification (IOI) 
task~\citep{wang2023interpretability}: given an input {\em John and Mary went to the park. John passed the bottle to}, the model should predict {\em Mary}. Further, corrupting the input by replacing {\em Mary} with {\em Anne} would result in the output changing to \emph{Anne}. Let \(\vx^{\text{Mary}}_{j}\) and \(\vx^{\text{Anne}}_{j}\) represent the original and corrupted residual streams at decoder block \(j\), depending on whether \emph{Mary} or \emph{Anne} was injected at the input.  Now, in a corrupted forward pass, for a given \(j\), if the replacement of \(\vx^{\text{Anne}}_{j}\) by \(\vx^{\text{Mary}}_{j}\) results in the restoration of the output token from {\em Anne} to {\em Mary}, then one can conclude that attention mechanism at decoder block \(j\) is responsible for moving the name information. This is an example of patching {\em corrupted-to-clean} activation, often called {\em causal tracing}~\citep{rome}. Activation patching, in general, refers to both clean-to-corrupted as well as corrupted-to-clean patching~\citep{zhang2023patching}. 

{\bf Knockout} is a method to prune nodes in the full computational graph of the model to identify task-specific circuits. Complete ablation of a node is equivalent to replacing the node output with an all-zero vector. Our initial experiments suggest that such an ablation destructively interferes with the model's computation. Instead, we follow~\citet{wang2023interpretability} for {\em mean-ablation} to perform knockouts. Specifically, we construct inputs from the false ontologies provided in the PrOntoQA dataset and compute the mean activations for each layer across different inputs:
\begin{align}
    \vx^{\text{Knock}}_{j} = \operatorname{Mean}_{i}\left (\{\vx^i_j|s_i\in S\in \mathcal{D}_\text{False}\}\right )
\end{align}
where \(\mathcal{D}_\text{False}\) denotes the false ontology dataset. Then, the language model function with head \(h_{j,k}\) knocked out for the residual stream corresponding to the \(l\)-th token, can be represented as,
\begin{align}
    s^\text{Knock}_i = \arg \max_{\vx^{i}_\text{logit}}\operatorname{LM}^l_{j,k}(\vx^{i}_\text{logit}|S, \vy^l_{j,k}=x^{\text{Knock}}_{j})
\end{align}

More often than not in this work, we will need to knock out a set of heads \(\mathcal{H}\); we will denote the corresponding language model as \(\operatorname{LM}^l_\mathcal{H}\). Also, if we perform knockout on the last residual stream (which, when projected to the token space, gives us the output token at the current step of generation), we will drop the superscript~\(l\).

\section{Task composition}
\label{sec:task-composition}
\begin{figure}[!t]
    \centering
\includegraphics[width=\textwidth]{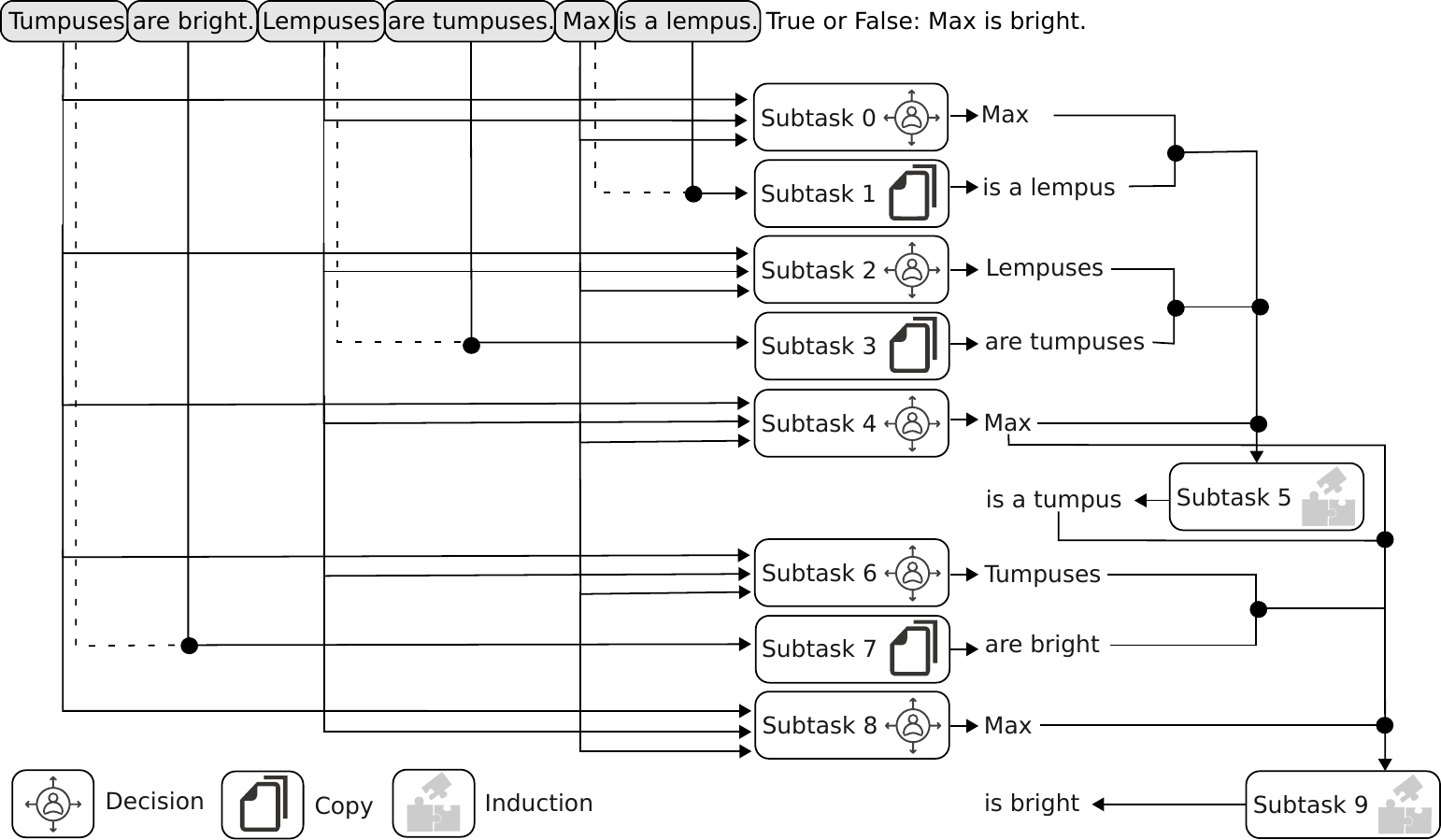}
    \caption{\color{black} {\bf An working example of task decomposition of CoT generation on fictional ontology}. {\em Decision}-type subtasks (0, 2, 4, 6, and 8) choose which reasoning path to follow; such decisions may be about the order of the information to be copied from the question to the generated output or whether inductive reasoning will be performed. {\em Copy}-type subtasks (1, 3, and 7) follow Decision subtasks and copy statements from the context to output. {\em Induction}-type subtasks (5 and 9) perform the reasoning steps of the form ``if A is B and B is C, then A is C''.}
    \label{fig:full-circuit-diagram}
\end{figure}
We seek to discover the circuits responsible in {\color{black} Llama-2} 7B for few-shot CoT reasoning on the examples from PrOntoQA fictional ontology problems. Consider the example presented in 
Figure~\ref{fig:full-circuit-diagram}: we ask the model 
``{\em Tumpuses are bright. Lempuses are tumpuses. Max is a lempus. True or False: Max is bright}'' (for brevity, we omit the few-shot examples that precede and the CoT prompt {\em Let's think step by step} that follows the question). {\color{black} Llama-2} 7B generates a verbose CoT reasoning sequence as follows: ``{\em Max is a lempus. Lempuses are tumpuses. Max is a tumpus. Tumpuses are bright. Max is bright.}'' We can observe that such a reasoning process constitutes three critical kinds of subtasks:
\begin{enumerate}[leftmargin=0.5cm]
    \item {\bf Decision-making}: The model decides on the path of reasoning to follow. In Figure~\ref{fig:full-circuit-diagram}, given the three possible entities to start with, namely {\em Tumpus}, {\em Lempus}, and {\em Max}, {\color{black} Llama-2} 7B starts with the last one (i.e., {\em Max is a lempus}). One would require multiple such decision-making steps within the complete reasoning process.
    \item {\bf Copying}: The LM needs to copy key information given in the input to the output. Typically, decision-making precedes copying, as the model needs to decide which information to copy.
    \item {\bf Induction}: The LM uses a set of statements to infer new relations. Again, a decision-making step precedes as the model must decide on the relation to infer. In the majority of this work, we focus on 2-hop inductions where given statements of the form {\tt A is B} and {\tt B is C}, the model should infer {\tt A is C}.
\end{enumerate}
%
{\color{black} In all our analyses henceforth, we follow a structure where the model needs to generate CoT responses that solve the overall task using ten steps or subtasks, each belonging to one of the three categories, as presented in Figure~\ref{fig:full-circuit-diagram}. (See Appendix~\ref{app:sec:prompts-and-model} for details of few-shot examples used along with the overall performance of {\color{black} Llama-2} 7B.) A natural hypothesis would be that distinct, well-defined circuits corresponding to each category of subtasks (i.e., decision-making, copying, and inductive reasoning) exist that work in conjunction to lead the CoT generation. However, as our findings in the next section suggest, this is not the case in reality.}

\subsection{Task-specific head identification}
\label{subsec:task-specific-heads}

As \cite{wang2023interpretability} suggested, the very first step toward circuit discovery is to identify the components. Since we presume the attention heads as the nodes of a circuit, the goal then becomes to identify those heads in the language model that are most important for a given task. We define the importance of the \(k\)-th head at the \(j\)-th decoder block, \(h_{j,k}\) for a particular task as follows. 

Let the task be defined as predicting a token \(s_{i+1}\) given the input \(S=\{s_1, s_2, \cdots, s_i\}\). For example, in the demonstration provided in Figure~\ref{fig:full-circuit-diagram}, predicting the token {\em Max} given the context {\em Tumpuses are bright. Lempuses are tumpuses. Max is a lempus. True or False: Max is bright. Response: Let us think step by step. Tumpuses are bright. Lempuses are tumpuses.} We assign a score \(\mu_\text{Task}(h_{j,k})\) to each head \(h_{j,k}\) proportional to their importance in performing the given task. Following the intuitive subtask demarcation presented earlier, we start with scoring the attention heads for each different subtask. We provide the detailed procedure of calculating \(\mu_\text{Task}(h_{j,k})\) for each category of subtasks --- decision-making, copying, and induction, in Appendix~\ref{app:task-specific-heads}.

\begin{figure}[!t]
    \centering
    \includegraphics[width=\textwidth]{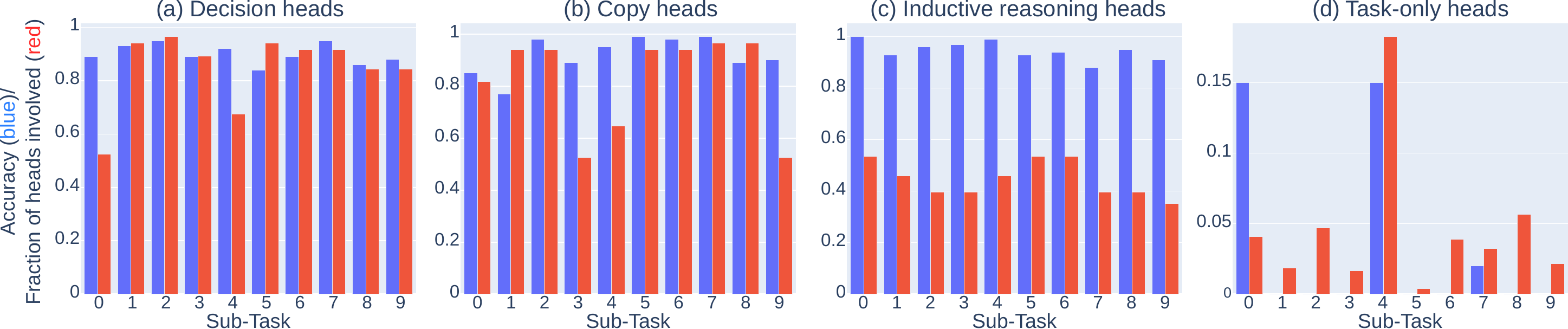}
    \caption{{\color{black}{\bf Performance of attention heads identified for each different subtask type across different subtasks.}  We show the performance of (a) decision-making, (b) copy, and (c) inductive reasoning heads for each subtask 0 to 9 (blue bars show accuracy when the rest of the heads are knocked out; red bars denote the fraction of heads involved, see Figure~\ref{fig:full-circuit-diagram} for subtask annotation). (d) Task-only heads are only those that are not shared with other tasks, e.g., only those copying heads for subtask 4 that are not decision-making or inductive reasoning heads. Inductive reasoning heads are consistently functional across all the subtasks with the least number of heads involved.}}
    \label{fig:task-wise-heads}
\end{figure}


\begin{figure}[!t]
    \includegraphics[width=\columnwidth]{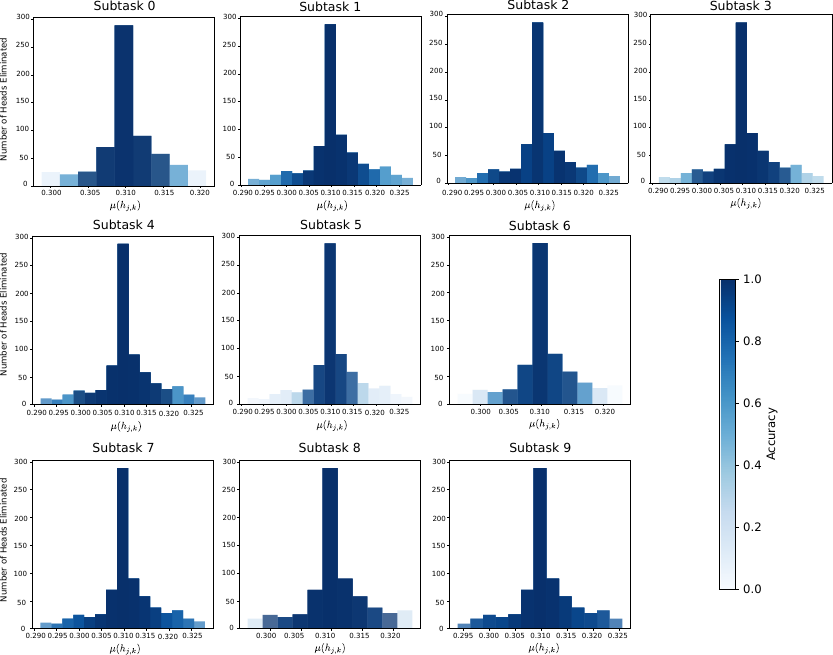}
    \caption{\color{black}{\bf Importance of inductive reasoning heads across subtasks}. We plot the counts of attention heads binned over relative head importance (\(\mu(h_{j,k})\)) computed on inductive reasoning task across all subtask indices (0-9, left to right and top to bottom). For a given subtask and a given bin of attention heads (corresponding to a range of \(\mu(h_{j,k})\)) knocked out, we show the accuracy of the pruned model model relative to the original model (color coded in blue). A wider spread of dark-colored bins for a subtask signifies that a higher number of attention heads can be knocked out without hurting the model performance; for example, in subtask 7, the removal of attention heads with \(\mu(h_{j,k})\in [0.295, 0.325]\) retains a similar performance compared to heads removed in subtask 5 with \(\mu(h_{j,k})\in [0.305, 0.315]\), signifying that more attention heads are required to remain active in subtask 5 compared to 7.}
    \label{fig:induction-all-index}
\end{figure}

Figure~\ref{fig:task-wise-heads} demonstrates how each of these three categories of attention heads performs on its respective subtasks and the rest of the subtasks (note that we perform head ablation here for each subtask index independently). We also show the fraction of the total number of heads involved in performing each subtask. A higher accuracy with a lower number of head involvement would suggest that we have found the minimal set of heads constitute the responsible circuit. As we can observe, there are no obvious mappings between the tasks used for head identification and the subtasks that constitute the CoT. {\color{black} Quite surprisingly, heads that we deem responsible for inductive reasoning can perform well across all the subtasks (Figure~\ref{fig:task-wise-heads} (c)) and requires the least number of active heads across all tasks}. Furthermore, the three sets of heads share a significant number of heads that are essential for the subtasks. In Figure~\ref{fig:task-wise-heads}(d), for each subtask, we use the heads that are not shared by subtasks of other categories. {\color{black} For example, in subtask 2, since we assumed it to be a decision-making subtask, we took only those decision-making heads that were not present in the copying or inductive reasoning heads}.
It is evident that these tasks are not structurally well differentiated in the language model. A good majority of heads share the importance of all three subtasks. The existence of backup circuits justifies such phenomena; however, with large models, the 
``self-repairing'' tendency~\citep{mcgrath2023hydra} is much higher. We hypothesize that there exists even further granularity of functional components that collectively express all three types of surface-level subtasks. We seek a solution in the earlier claims that pattern matching via induction heads serves as the progenitor of in-context learning~\citep{olsson2022icl-induction-heads}. We observe that the inductive reasoning subtasks of the form {\tt if [A] is [B] and [B] is [C] then [A] is [C]} essentially requires circuits that can perform 1) representation mixing from {\tt [A]} to {\tt [B]} to {\tt [C]}, and 2) pattern matching over past context by {\em deciding} which token to copy and then copying them to the output residual stream. Therefore, heads responsible for inductive reasoning can perform decision-making and copying by this argument. 

For empirical validation of this argument, we analyze the effects of knocking out heads on the accuracy of each of the subtask indices. For each subtask, we construct a histogram of head counts over \(\mu(h_{j,k})\). It can be seen that the corresponding density distribution is Gaussian in nature. Furthermore, we observe that the actual effect of pruning a head on the accuracy is not dependent on the actual value of \(\mu(h_{j,k})\) but varies with \(\delta(h_{j,k})=(\mu(h_{j,k}) - \operatorname{Mean}_{j,k}\left(\mu(h_{j,k})\right)^2\). From the definition of \(\mu(h_{j,k})\) provided in Appendix~\ref{app:task-specific-heads}, we can explain the behavior as follows: a very low value of \(\mu(h_{j,k})\) is possible when the KL-divergence between the noisy logits and original logits is almost same as the KL-divergence between patched and original logits. Conversely, \(\mu(h_{j,k})\) can be very high when the KL-divergence between the patched logit and the original logit is very small. Both of these suggest a strong contribution of the particular head.

{\color{black}Figure ~\ref{fig:induction-all-index} shows the accuracy of the model at each different subtask indices with heads pruned at different \(\delta(h_{j,k})\) (we show the histogram over \(\mu(h_{j,k})\) and start pruning heads from the central bin corresponding to the smallest values of \(\delta(h_{j,k})\)). Among the $1,024$ heads of {\color{black} Llama-2} 7B, we can see that pruning the model to as low as $\sim$400 heads retains \(>\)90\% of the original accuracy (see Table~\ref{tab:head-removal} in Appendix for subtask-wise statistics).} 

Further patterns in allocating attention heads for different subtasks can be observed in Figure~\ref{fig:induction-all-index}. Typically, subtask indices 0 and 5 are most sensitive to pruning attention heads; the accuracy of the model quickly decreases as we knock out attention heads with \(\mu(h_{j,k})\) deviated from the mean value. Subtask-0 is a decision-making subtask where the model decides on the reasoning path. One can further demarcate between subtasks 0 vs 2, 6, and 8, while we initially assumed all of these four to be decision-making stages. Referring to the example provided in Figure~\ref{fig:full-circuit-diagram}, in subtask-0, the model has very little information about which ontological statement to choose next, compared to subtask indices 2, 6, and 8, where the model needs to search for the fact related to the second entity in the last generated fact. Subtask 5 is the actual inductive reasoning step. Compared to subtask 9, which is also an inductive reasoning step, subtask 5 does not have any prior (the answer generated in subtask 9 is framed as a question in the input).

\section{Token mixing over fictional ontology}
\label{sec:token-mixing}

{\color{black} In the earlier section, we observed that the set of heads that we identify as responsible for inductive reasoning performs consistently across all subtasks with the least number of heads remaining active, retaining more than 90\% of the original performance with as low as 40\% of the total number of heads.
We continue to investigate the inductive reasoning functionality with this reduced set of heads.} Recall that we first need to explore if there is indeed a mixing of information happening across tokens of the form {\tt [A] is [B]}. Specifically, we pose the following question -- given the three pairs of tokens ({\tt [A\textsubscript{1}]}, {\tt [B\textsubscript{1}]}), ({\tt [A\textsubscript{2}]}, {\tt [B\textsubscript{2}]}) and ({\tt [A\textsubscript{3}]}, {\tt [B\textsubscript{3}]}), such that there exists a positive relation {\tt [A\textsubscript{1}] is [B\textsubscript{1}]}, a negative relation  {\tt [A\textsubscript{2}] is not [B\textsubscript{2}]}, and no relations exist between {\tt [A\textsubscript{3}]} and {\tt [B\textsubscript{3}]}, is it possible to distinguish between \([\vx^{A_1}_j: \vx^{B_1}_j]\), \([\vx^{A_2}_j: \vx^{B_2}_j]\), and \([\vx^{A_3}_j: \vx^{B_3}_j]\), where \([\cdot:\cdot]\) denotes the concatenation operator, for a given decoder layer \(j\)? Given that these tokens denote entities from a fictional ontology, the model could not possibly have memorized similar representations with the embedding/MLP blocks for the tokens. Instead, it needs to deploy the attention heads to move information from the residual stream of one token. 

We translate the posed question into learning a probing classifier~\citep{2022-probing} \(\mathcal{C}:\mathcal{X}\rightarrow \mathcal{Y}\), where for each \([\vx^{A_i}_j: \vx^{B_i}_j]\in \mathcal{X}\), the corresponding label \(y\in \mathcal{Y}=\{-1, 0, +1\}^{\lvert\mathcal{X}\rvert}\) should be \(-1\), \(0\), or \(+1\), if {\tt A\textsubscript{i}} and {\tt B\textsubscript{i}} are negatively related, unrelated, or positively related, respectively. {\color{black} There can be multiple occurrences of a token in the input context (e.g., there are two occurrences of [B] in the above example). As a result, there will be multiple residual streams emanating from the same token. To decide which residual streams to pair with (in positive/negative pairing), we rely on immediate occurrence. For example, given the input {\tt [A] is [B]. [A] is not [C]}, we take the first occurrence of [A] to be positively related to [B], while the second occurrence of [A] as negatively related to [C].} We experiment with a linear and a non-linear implementation of \(\mathcal{C}\) as feedforward networks. Specifically, in the linear setting, we use a single linear transformation of the form \(y = (\mW\vx + \mB)\) followed by a softmax; in the nonlinear setup, we utilize stacked linear transformations with ReLU as intermediate non-linearity. We provide the full implementation details of this experiment in Appendix~\ref{app:sec:probing}. Given that the baseline distribution of the residual stream representation might vary across layers, we employ layer-specific classifiers.

\begin{figure}[!t]
    \centering
    \includegraphics[width=\columnwidth]{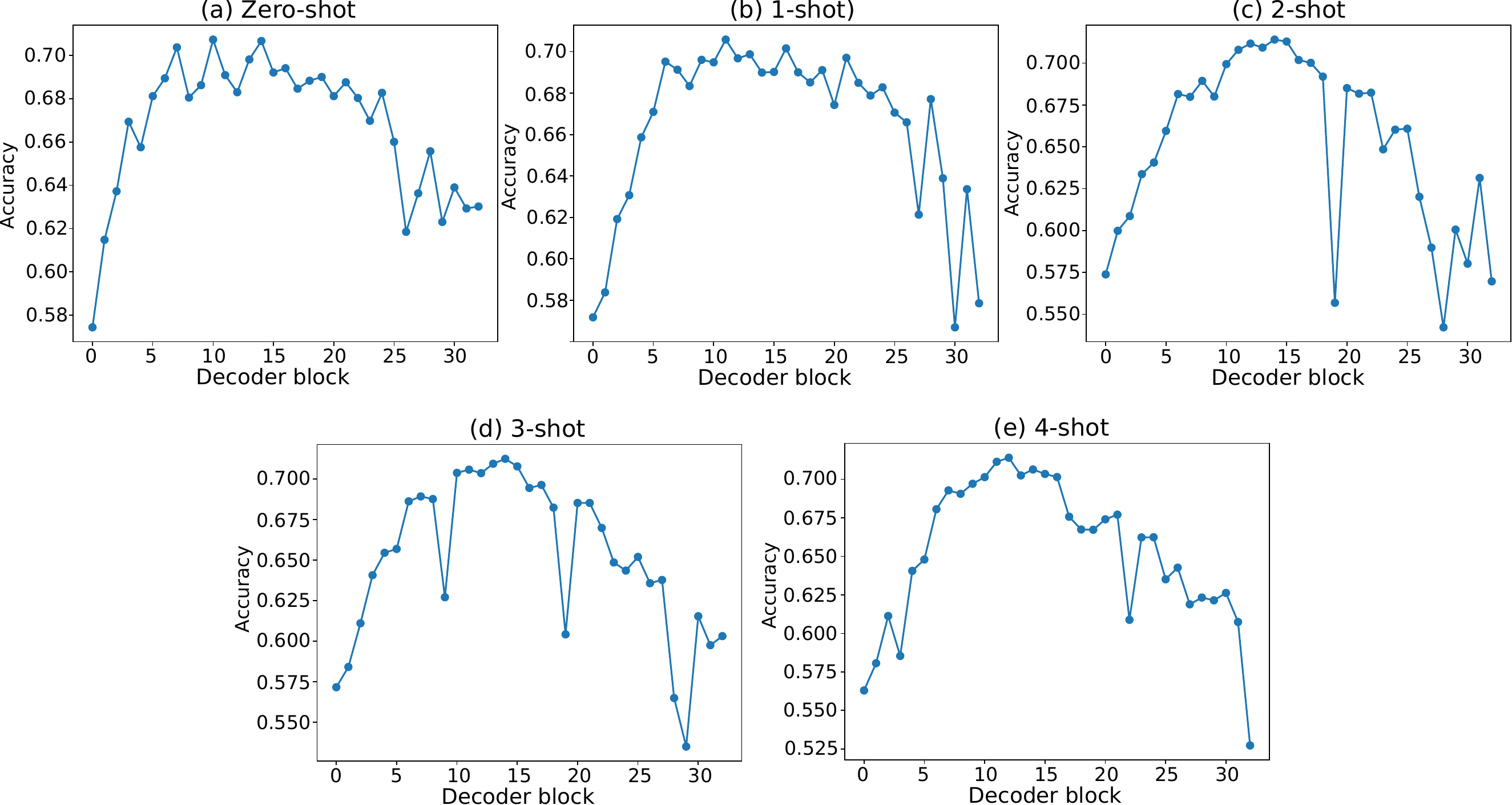}
    \caption{{\bf How does the LLM mix information among tokens according to ontology?} We plot the performance (in terms of accuracy) of classifying whether two tokens are ontologically related, unrelated, or negatively related using their residual stream representations at different layers. We demonstrate the depthwise accuracy profile using different numbers of few-shot examples (0 to 4). Typically, mixing information between tokens according to their relation does not require in-context learning.  Information mixing between related (or negatively related) token pairs results in better distinguishability: gradually increasing from the starting decoder blocks and achieving a peak between decoder blocks 10-15.}
    \label{fig:probing-result}
\end{figure}

Figure~\ref{fig:probing-result} shows the 3-way classification performance in terms of accuracy across different layers using 4-layer ReLU networks. The following conclusions can be drawn from the observations:
\begin{itemize}[leftmargin=0.5cm]
    \item {\bf Non-linear information movement between residual streams.} Our experiments with the linear classifier fail to distinguish between ontologically related, unrelated, and negatively related entities.
    
    \item {\bf Distinguishability of the residual stream pairs improves across depth.} While the very first layer of attention provides a substantial degree of required token mixing, we observe that the classification performance goes better gradually, pointing towards the existence of multiple successive attention heads that continue moving information from {\tt A\textsubscript{j}} to {\tt B\textsubscript{j}} if they are related. However, after a certain depth, the distinguishability starts diminishing, most likely due to accumulation of other information related to the tasks.

    \item {\bf Token mixing is not boosted from contextual prior.} Across different numbers of in-context examples provided, we do not observe any significant difference in peak classification performance compared to zero-shot regimes. However, as the number of examples increases, we observe unstable classification performance. It is likely that with more in-context examples, task-specific information is accumulated earlier than the zero-shot regime.
\end{itemize}

\section{Circuitry of step-by-step generation}
\label{sec:circuit-I}

In Section~\ref{subsec:task-specific-heads}, we observed that the head importance identified via inductive reasoning task could serve as a proxy to identify the important heads across all the subtasks. This provides us with an opportunity to reduce the total number of attention heads to analyze when investigating the step-by-step generation procedure. For each subtask, we use a threshold range for \(\mu_{h,k}\)  to select a subset of attention heads that are kept intact while the rest of the heads are knocked out (see Appendix~\ref{app:task-specific-heads} for the subtask-wise threshold range used and the corresponding subtask accuracy). This chosen subset of the model gives us an aggregate accuracy of \(0.9\) over the inputs for which the full model generates correctly.

\begin{figure}[!t]
    \centering
    \includegraphics[width=\columnwidth]{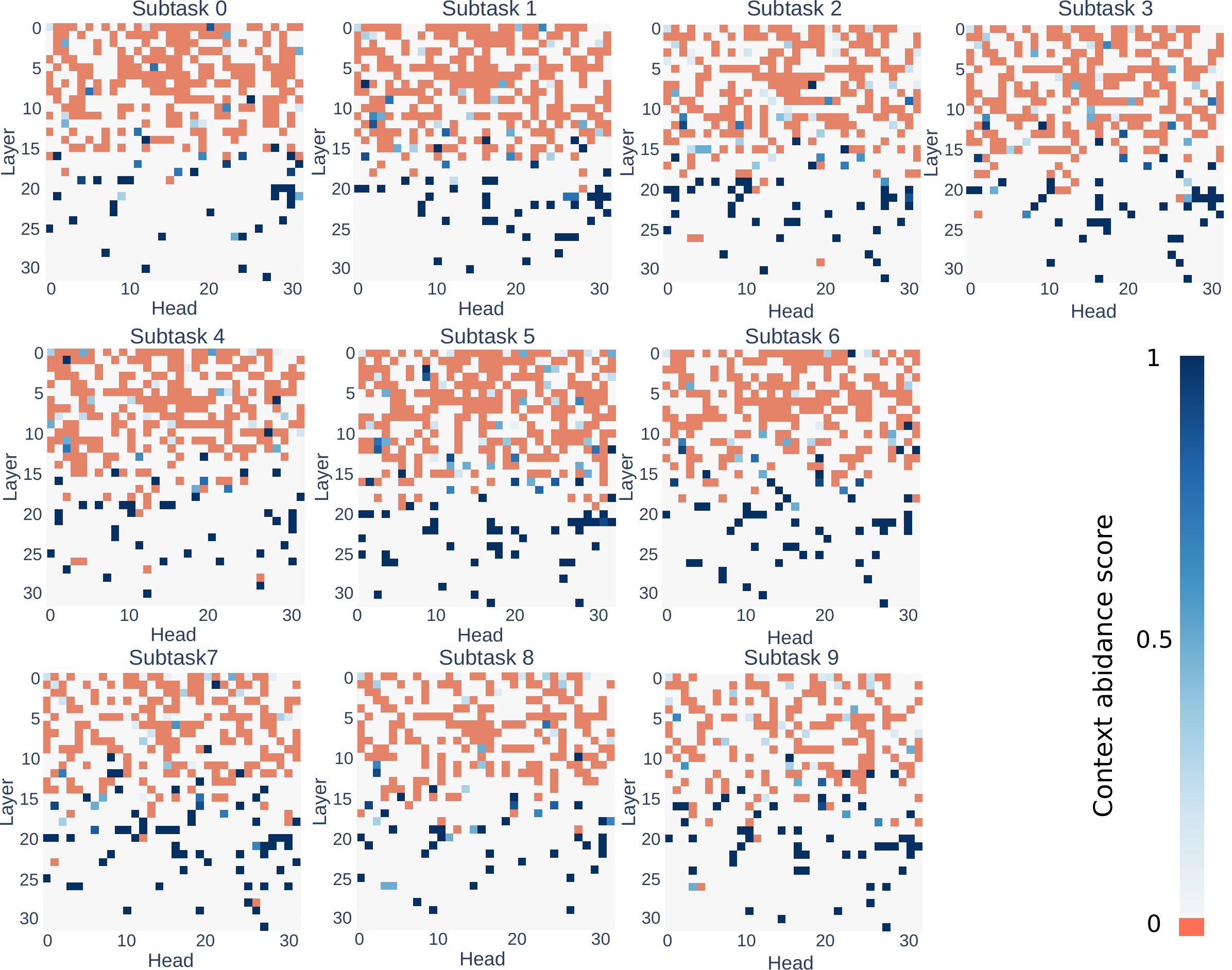}
    \caption{{\bf When does the LLM start following the context?} Distribution of context-abidance score \(c_{j,k}\) of each head (darker blue shade signifies higher \(c_{j,k}\)) with layer (\(j\)) on y-axis and head index (\(k\)) on x-axis. Attention heads in red denote those with zero context-abidance. We show the distributions for all the subtask indices 0-9 (left to right, top to bottom). Typically, context abidance is task-agnostic and emerges after the 16th decoder block.}
    \label{fig:context-dependence}
\end{figure}
\begin{figure}[!t]
    \centering
    \includegraphics[width=\textwidth]{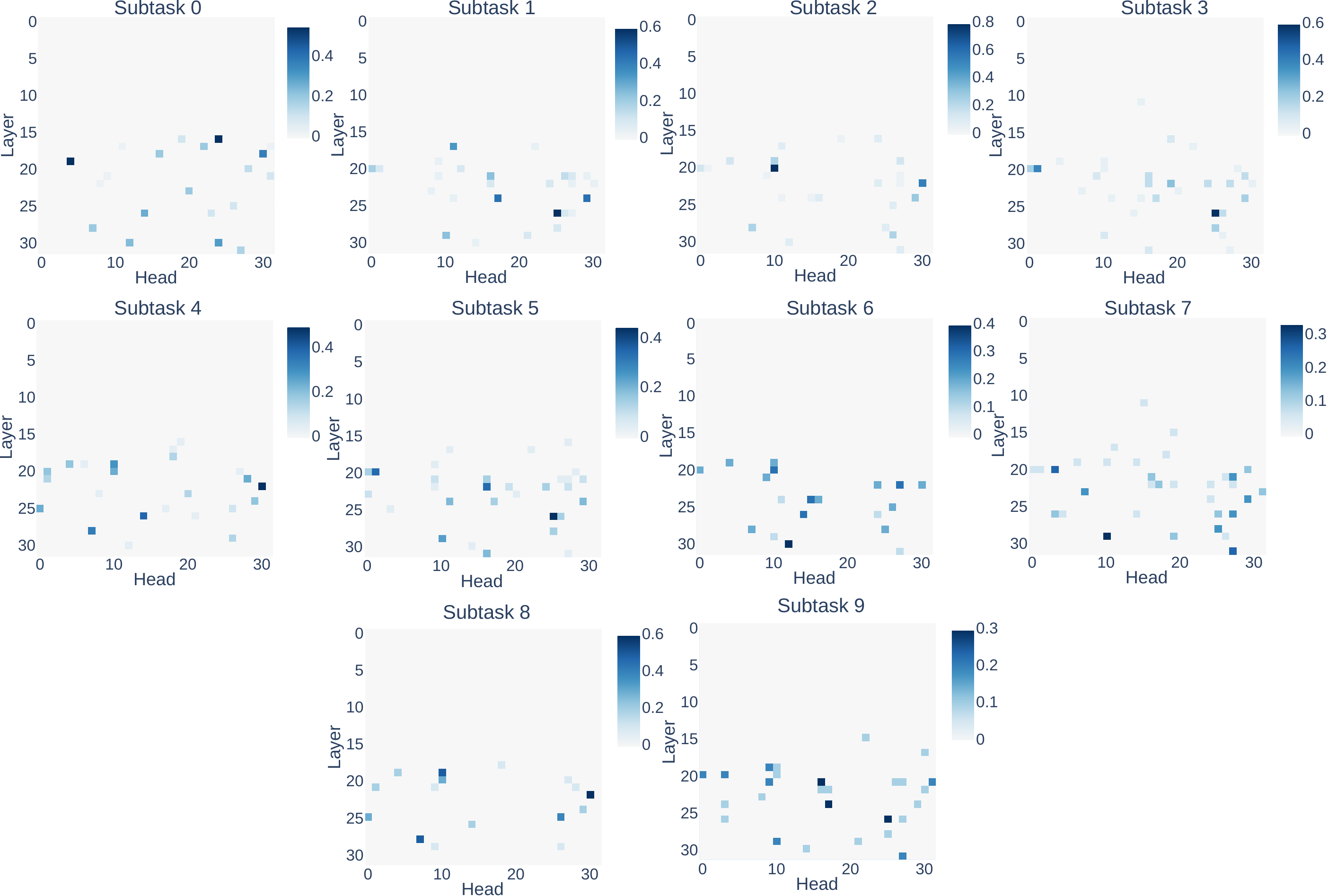}
    \caption{{\bf Which heads are writing the answer?} For each subtask (0-9), we show the attention heads that write the answer to the corresponding task into the last residual stream, along with the probability of the answer token in the attention head output.}
    \label{fig:ans-writing-head}
\end{figure}

Next, we proceed to look into the exact information that is being read and written by the attention heads from and into the residual streams. Given a head \(h_{j,k}\) that writes \(\vy^i_{j,k}\) to the residual stream \(\vx^i_{j-1}\), we apply the unembedding projection \(\mU\) on \(\vy^i_{j,k}\) and select the token with the highest probability, \(\hat{s}^{j,k}_i\). The unembedding projection provides the opportunity to look into the token-space representation directly associated with the residual stream and its subspaces~\citep{logit-lens}. However, it should be noted that applying unembedding projection typically corresponds to Bigram modeling~\citep{elhage2021mathematical}. Therefore, when we map any intermediate representation (attention output, residual stream, etc.), we essentially retrieve the token that is most likely to follow.

\subsection{In-context prior vs pretraining prior}
\label{subsec:in-context}

We start by exploring the depth at which the model starts following the context provided as input. Specifically, we check for a given token \(s_i\) in the sequence \(S\), if \(\hat{s}^{j,k}_i\), the token projected by the attention head \(h_{j,k}\) is such that \(\langle s_i, \hat{s}^{j,k}_i \rangle\) is bigram present in \(S\). We compute a {\em context-abidance score}, \(c_{j,k}\) for the head \(h_{j,k}\) as the fraction of tokens for which the above condition holds true. 

In Figure~\ref{fig:context-dependence}, we plot the context abidance distribution for each head across different subtask indices. We can observe a visible correlation between the depth of the head and how much context abiding it is --- {\em the LM starts focusing on the contextual information at deeper layers}. Given that our experiment setup is based on fictional ontology, \(c_{j,k}\) provides a strong demarcation between pretraining prior (i.e., language model statistics memorized from pretraining) and contextual prior (i.e., language model statistics inferred from context) since there are negligible chances of the LM to memorize any language modeling statistics containing the fictional entities from the pretraining data. Therefore, it predicts a correct bigram only when it is able to follow the context properly. However, there is no visible demarcation across different subtasks; we can claim that this {\em depth-dependence is implicit to the model and does not depend on the prediction task}.

\subsection{Subtask-wise answer generation}
\label{subsec:answer-writer}

We investigate the information propagation pathway through the attention heads that constitute the step-by-step answering of the subtasks. We start with the heads that directly write the answer to the particular subtask into the last residual stream by mapping their output to the token space using \(\mU\). Figure~\ref{fig:ans-writing-head} plots these answer-writing heads along with the probability of the answer in the attention head output across different subtasks. The existence of multiple answer-writing heads suggests that LMs exploit multiple pathways to generate the same answer, with each such pathway reinforcing the generated output. Looking at the heads with the highest answer probability, we can observe that some top heads are utilized across different subtasks, e.g., subtasks 1 and 3, subtasks 2, 4, and 8, subtasks 5 and 9, etc.

{\bf Sharp change in depth-wise functionality.} Interestingly, the 16-th layer appears as a {\em region of functional transition} in the model with all the answer-writing heads appearing {\em after} this particular layer. Our earlier experiments also associate this layer with some phase shifts; in the token-mixing analysis (Section~\ref{sec:token-mixing}), the distinguishability of ontologically related entities is observed to achieve peaks near this layer and starts falling after that; in Section~\ref{subsec:in-context} as well, we see a certain overall rise in context-abidance in the heads after the 16-th layer (see Figure~\ref{fig:context-dependence}). All three findings suggest a {\em functional rift} in LLMs that happens to lie almost at the halfway point from embedding to unembedding. The embedding layer associates tokens with information available from pretraining. The initial half assists information movement between residual streams and {\em aligns} the representations to the contextual prior. The latter half of the model employs multiple pathways to write the answer to the last residual stream.
\begin{figure}[!t]
    \centering
    \includegraphics[width=\textwidth]{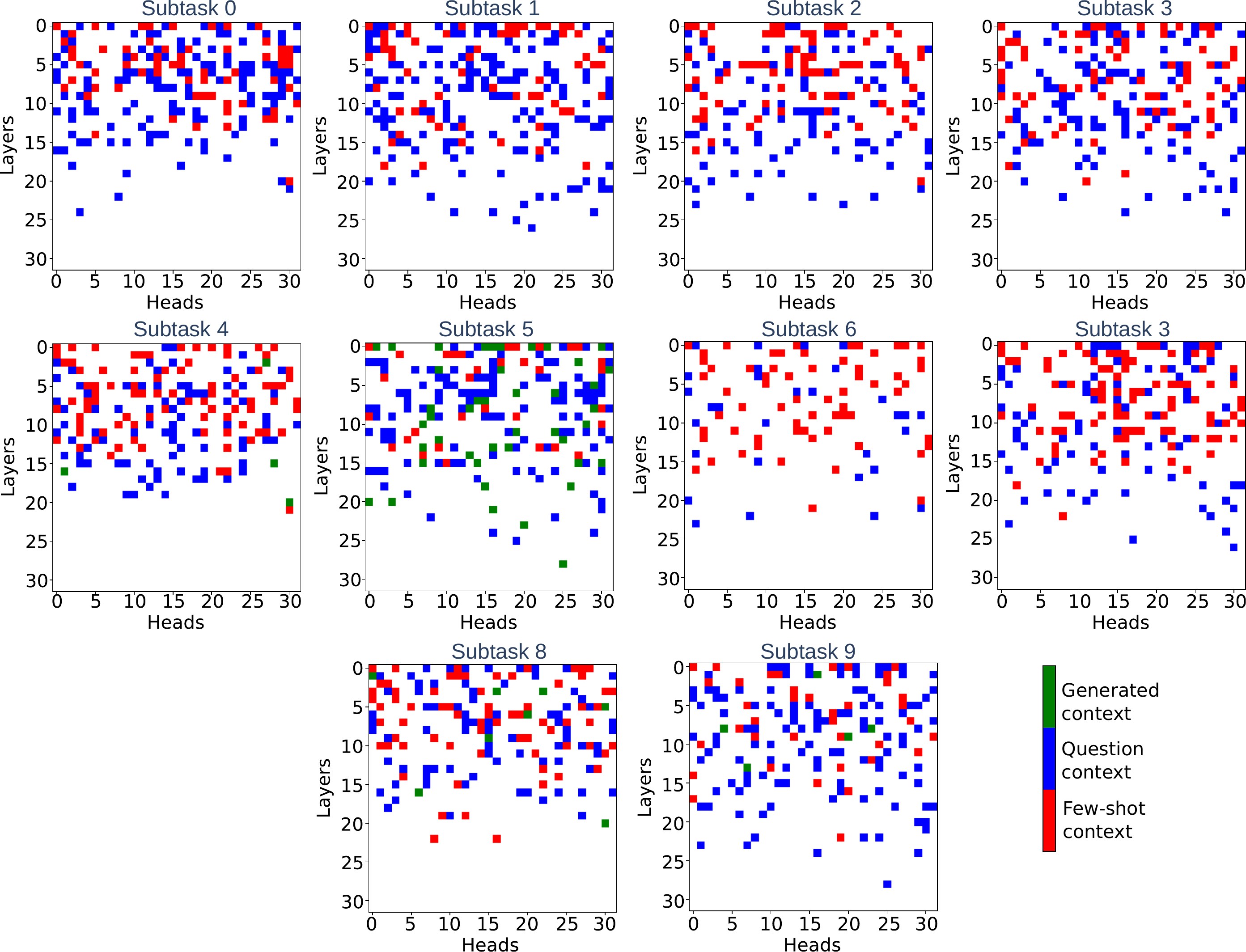}
    \caption{{\bf Where do the answer-writing heads collect their answers from?} For each answer-writing head and each subtask, we show those attention heads that attend to residual streams corresponding to answer tokens. We demarcate heads that collect the answer tokens from the generated context (green), question context (blue), and few-shot context (red).}
    \label{fig:answer-source}
\end{figure}

\subsection{Parallel pathways of answer processing}
\label{subsec:answer-collection}

Once we observe the existence of multiple answer-writers within the model, the natural step forward is to wonder if they all process the answer from the input using the same mechanism. We employ a recursive strategy to identify the flow of information through the attention heads (see Appendix~\ref{app:sec:info-flow} for a detailed description of the procedure). Specifically, we start from the answer-writing heads, follow which residual streams are being attended by these heads, identify the content of these residual streams via unembedding projection, and identify the heads in the previous layers that are writing that content into those residual streams. We continue till one of the two conditions is met: (i) we reach a head in the first decoder block, or (ii) we reach a residual stream corresponding to the first token in the input token sequence.
With such a procedure, we construct trees of attention heads rooted at the answer writing heads.

Now, let us consider the following example context plus generated token sequence: {\em Lempuses are tumpuses. Tumpuses are rompuses. Rompuses are blue. Max is lempus. True or false: Max is blue. Response: Let us think step by step. Max is lempus. Lempuses are tumpuses. Max is} with the desired prediction being {\em tumpus} (note that this input follows the few shot examples that are not shown here for brevity). We segregate the input context into three parts: few-shot context, question context, and generated context. We can see that the LLM can collect the answer {\em tumpus} from either generated or question context. Also, if the same token exists in the few-shot context, it can decide to collect information from there as well (however, that should not be ideal since the contextual role of that token would be different). We proceed to identify from the information-flow trees all those heads that attend to the streams corresponding to the answer token (in this example, {\em tumpus}). Note that there can be multiple other heads that might be attending to these same tokens; but only those heads that are elements of the tree rooted at answer writing heads are actually contributing to the flow of information from the answer tokens in the context of the output
Figure~\ref{fig:answer-source} demonstrates these attention heads that attend to the residual streams corresponding to the answer tokens for different subtasks, present in the generated context (green), question context (blue) and few-shot context (red). The following observations can be drawn from the figures: 
\begin{enumerate}[leftmargin=0.5cm]
    \item {\bf Coexitent pathways of answer generation}. Smaller models like GPT-2 on simpler tasks like IOI implements a unique neural algorithm~\citep{wang2023interpretability}. On the contrary, a larger model like {\color{black} Llama-2} exploits different parallel pathways of answer propagation while performing reasoning
    There are different heads that are directly connected to the answer writers, and they collect answer information from different places in the context.

    \item {\bf Different primary sources of answer for different subtasks.} In the initial stages of generation (i.e., subtasks 0, 1, 2, 3), the answer tokens are not present in the generated context. For subtasks 6 and 7, again, the answers need to be collected from the question index. For subtasks 4, 5, 8, and 9, we can observe the presence of heads that collect answer tokens from the context generated via earlier subtasks (in fact, subtask 5 has more such heads). 
\end{enumerate}

Given the fact that (i) there are multiple answer writing heads, and (ii) there are different heads that are directly connected to the answer writers, and they collect answer information from different places in the context, we can conclude that the different pathways implement different algorithms as well. In the few-shot examples, there are fictional entities that have been used in the question context as well, though in different contextual roles. The fact that there are attention heads collecting those entities as answers suggests that there are pathways prone to collecting similar information from the few-shot context. Although these pathways collect the same token as the answer, they actually deviate from the correct reasoning algorithm. However, the presence of such pathways decreases as the generation progresses from subtask 0 to 9. Note that this is different from few-shot examples providing the necessary pattern via in-context learning. Such patterns are more position-specific, i.e., while generating the answer for a certain subtask in the question, there is an overall increased attention provided to the same subtask tokens in the few-shot examples (see Figures~\ref{fig:attn-Input}, \ref{fig:few-shot-att-1-2}, \ref{fig:few-shot-att-3-4}, and \ref{fig:few-shot-att-5-6} in Appendix). Finally, the pattern of answer collection from question context and generated context across different subtasks clearly demarcates the inductive reasoning subtasks (5 and 9) from the decision-making and copying. Tasks in the former category require pathways that collect answers from the earlier generated context, but the latter does not. Moreover, the decision-making step right before inductive reasoning can extract answers from the preceding step. This explains an albeit small number of heads that collect the answer from the generated context in subtasks 4 and 8. See Figures~\ref{fig:InformationFlow} and \ref{fig:InformationFlow_noise_index_5} for examples of such information flow towards subtasks 1 and 5, respectively.


\section{Conclusion}
\label{sec:conclude}

{\bf Our findings.} This work sought to disentangle the functional fabric of CoT reasoning in LLMs. Specifically, we explored {\color{black} Llama-2} 7B in a few-shot CoT regime for solving multi-step reasoning problems on fictional ontologies of the PrOntoQA dataset. We observed that:
\begin{enumerate}[leftmargin=0.5cm]
    \item Despite different reasoning requirements across different stages of CoT generation, the functional components of the model remain almost the same. Different neural algorithms are implemented as compositions of induction circuit-like mechanisms.
    \item Attention heads perform information movement between ontologically related (or negatively related) tokens. This information movement results in distinctly identifiable representations for such token pairs. Typically, this distinctive information movement starts from the very first layer and continues till the middle. While this phenomenon happens zero-shot, in-context examples exert pressure to quickly mix other task-specific information among tokens.
    \item Multiple different neural pathways are deployed to compute the answer, that too in parallel. Different attention heads, albeit with different probabilistic certainty, write the answer token (for each CoT subtask) to the last residual stream. 
    \item These parallel answer generation pathways collect answers from different segments of the input. We found that while generating CoT, the model gathers answer tokens from the generated context, the question context, as well as the few-shot context. This provides a strong empirical answer to the open problem of whether LLMs actually use the context generated via CoT while answering questions.
    \item We observe a functional rift at the very middle of the LLM (16th decoder block in case of {\color{black} Llama-2} 7B), which marks a phase shift in the content of residual streams and the functionality of the attention heads. Prior to this rift, the model primarily assigns bigram associations memorized via pretraining; it drastically starts following the in-context prior to and after the rift. It is likely that this is directly related to the token-mixing along ontological relatedness that happens only prior to the rift. Similarly, answer-writing heads appear only after the rift. Attention heads that (wrongly) collect the answer token from the few-shot examples are also bounded by the prior half of the model.
\end{enumerate}

{\bf Implications for future research.} These findings bear important ramifications towards the ongoing research around language modeling and interpretability. A natural extension to this work would be to incorporate pretraining memorization in terms of MLP blocks --- precisely, whether the functional rift across layers bears a similar mechanism when the LLM starts mixing factual associations that have been stored within the MLP neurons. The existence of a parallel answer-generation process is extremely important for causal interventions on model behavior~\citep{li2024circuit-breaking}: changing how a model should reason via up-(or down-) scaling certain neural pathways should take all the parallel pathways into account.

{\color{black}{\bf Limitations.} The design of this study imposes certain limitations in its scope. First and foremost, we analyze a very specific type of reasoning problem. While the fictional ontology with a restricted CoT template provides ease of analysis, free-form reasoning can introduce further complex dynamics not captured in this study. We also could not address the role of MLPs in reasoning. While existing literature points to their role as factual memory, it should be noted that the model also memorizes a diverse set of token-token associations pertaining to the structure of language and not just factual associations. Such associations are likely to play a decisive role in a grounded reasoning setup that our analysis ignores. Finally, we use a few-shot prompting regime to ensure that the model follows a specific structure of reasoning steps. With zero-shot CoT, more complex mechanisms are bound to emerge.
}

\bibliography{tmlr}

\begin{thebibliography}{31}
\providecommand{\natexlab}[1]{#1}
\providecommand{\url}[1]{\texttt{#1}}
\expandafter\ifx\csname urlstyle\endcsname\relax
  \providecommand{\doi}[1]{doi: #1}\else
  \providecommand{\doi}{doi: \begingroup \urlstyle{rm}\Url}\fi

\bibitem[Belinkov(2022)]{2022-probing}
Yonatan Belinkov.
\newblock {Probing Classifiers: Promises, Shortcomings, and Advances}.
\newblock \emph{Computational Linguistics}, 48\penalty0 (1):\penalty0 207--219, 04 2022.
\newblock ISSN 0891-2017.
\newblock \doi{10.1162/coli_a_00422}.
\newblock URL \url{https://doi.org/10.1162/coli\_a\_00422}.

\bibitem[Bi et~al.(2024)Bi, Zhang, Jiang, Deng, Zheng, and Chen]{bi2024program}
Zhen Bi, Ningyu Zhang, Yinuo Jiang, Shumin Deng, Guozhou Zheng, and Huajun Chen.
\newblock When do program-of-thought works for reasoning?, 2024.

\bibitem[Chen et~al.(2023)Chen, Ma, Wang, and Cohen]{chen2023pot}
Wenhu Chen, Xueguang Ma, Xinyi Wang, and William~W. Cohen.
\newblock Program of thoughts prompting: Disentangling computation from reasoning for numerical reasoning tasks.
\newblock \emph{Transactions on Machine Learning Research}, 2023.
\newblock ISSN 2835-8856.
\newblock URL \url{https://openreview.net/forum?id=YfZ4ZPt8zd}.

\bibitem[Elhage et~al.(2021)Elhage, Nanda, Olsson, Henighan, Joseph, Mann, Askell, Bai, Chen, Conerly, et~al.]{elhage2021mathematical}
Nelson Elhage, Neel Nanda, Catherine Olsson, Tom Henighan, Nicholas Joseph, Ben Mann, Amanda Askell, Yuntao Bai, Anna Chen, Tom Conerly, et~al.
\newblock A mathematical framework for transformer circuits.
\newblock \emph{Transformer Circuits Thread}, 1, 2021.

\bibitem[Elhage et~al.(2022)Elhage, Hume, Olsson, Schiefer, Henighan, Kravec, Hatfield-Dodds, Lasenby, Drain, Chen, et~al.]{elhage2022toy-superposition}
Nelson Elhage, Tristan Hume, Catherine Olsson, Nicholas Schiefer, Tom Henighan, Shauna Kravec, Zac Hatfield-Dodds, Robert Lasenby, Dawn Drain, Carol Chen, et~al.
\newblock Toy models of superposition.
\newblock \emph{arXiv preprint arXiv:2209.10652}, 2022.

\bibitem[Feng et~al.(2023)Feng, Gu, Zhang, Ye, He, and Wang]{feng2023CoT-with-circuits}
Guhao Feng, Yuntian Gu, Bohang Zhang, Haotian Ye, Di~He, and Liwei Wang.
\newblock Towards revealing the mystery behind chain of thought: a theoretical perspective.
\newblock \emph{arXiv preprint arXiv:2305.15408}, 2023.

\bibitem[Geva et~al.(2021)Geva, Schuster, Berant, and Levy]{geva-etal-2021-transformer}
Mor Geva, Roei Schuster, Jonathan Berant, and Omer Levy.
\newblock Transformer feed-forward layers are key-value memories.
\newblock In \emph{Proceedings of the 2021 Conference on Empirical Methods in Natural Language Processing}, pp.\  5484--5495. Association for Computational Linguistics, 2021.
\newblock \doi{10.18653/v1/2021.emnlp-main.446}.
\newblock URL \url{https://aclanthology.org/2021.emnlp-main.446}.

\bibitem[Henighan et~al.(2023)Henighan, Carter, Hume, Elhage, Lasenby, Fort, Schiefer, and Olah]{henighan2023superposition}
Tom Henighan, Shan Carter, Tristan Hume, Nelson Elhage, Robert Lasenby, Stanislav Fort, Nicholas Schiefer, and Christopher Olah.
\newblock Superposition, memorization, and double descent.
\newblock \emph{Transformer Circuits Thread}, 2023.

\bibitem[Kojima et~al.(2022)Kojima, Gu, Reid, Matsuo, and Iwasawa]{AutoCoT}
Takeshi Kojima, Shixiang~(Shane) Gu, Machel Reid, Yutaka Matsuo, and Yusuke Iwasawa.
\newblock Large language models are zero-shot reasoners.
\newblock In S.~Koyejo, S.~Mohamed, A.~Agarwal, D.~Belgrave, K.~Cho, and A.~Oh (eds.), \emph{Advances in Neural Information Processing Systems}, volume~35, pp.\  22199--22213. Curran Associates, Inc., 2022.
\newblock URL \url{https://proceedings.neurips.cc/paper_files/paper/2022/file/8bb0d291acd4acf06ef112099c16f326-Paper-Conference.pdf}.

\bibitem[Lampinen et~al.(2022)Lampinen, Dasgupta, Chan, Mathewson, Tessler, Creswell, McClelland, Wang, and Hill]{lampinen-etal-2022-CoT-no-causal}
Andrew Lampinen, Ishita Dasgupta, Stephanie Chan, Kory Mathewson, Mh~Tessler, Antonia Creswell, James McClelland, Jane Wang, and Felix Hill.
\newblock Can language models learn from explanations in context?
\newblock In \emph{Findings of the Association for Computational Linguistics: EMNLP 2022}, pp.\  537--563. Association for Computational Linguistics, 2022.
\newblock \doi{10.18653/v1/2022.findings-emnlp.38}.

\bibitem[Li et~al.(2024)Li, Davies, and Nadeau]{li2024circuit-breaking}
Maximilian Li, Xander Davies, and Max Nadeau.
\newblock Circuit breaking: Removing model behaviors with targeted ablation, 2024.

\bibitem[Liu et~al.(2022)Liu, Ash, Goel, Krishnamurthy, and Zhang]{liu2022shortcut-automata}
Bingbin Liu, Jordan~T Ash, Surbhi Goel, Akshay Krishnamurthy, and Cyril Zhang.
\newblock Transformers learn shortcuts to automata.
\newblock \emph{arXiv preprint arXiv:2210.10749}, 2022.

\bibitem[McGrath et~al.(2023)McGrath, Rahtz, Kramar, Mikulik, and Legg]{mcgrath2023hydra}
Thomas McGrath, Matthew Rahtz, Janos Kramar, Vladimir Mikulik, and Shane Legg.
\newblock The hydra effect: Emergent self-repair in language model computations.
\newblock \emph{arXiv preprint arXiv:2307.15771}, 2023.

\bibitem[Meng et~al.(2022)Meng, Bau, Andonian, and Belinkov]{rome}
Kevin Meng, David Bau, Alex Andonian, and Yonatan Belinkov.
\newblock Locating and editing factual associations in gpt.
\newblock In S.~Koyejo, S.~Mohamed, A.~Agarwal, D.~Belgrave, K.~Cho, and A.~Oh (eds.), \emph{Advances in Neural Information Processing Systems}, volume~35, pp.\  17359--17372. Curran Associates, Inc., 2022.
\newblock URL \url{https://proceedings.neurips.cc/paper_files/paper/2022/file/6f1d43d5a82a37e89b0665b33bf3a182-Paper-Conference.pdf}.

\bibitem[Nanda(2022)]{nanda2022comprehensive}
Neel Nanda.
\newblock A comprehensive mechanistic interpretability explainer \& glossary, 2022.

\bibitem[nostalgebraist(2020)]{logit-lens}
nostalgebraist.
\newblock interpreting {G}{P}{T}: the logit lens — {L}ess{W}rong --- lesswrong.com.
\newblock \url{https://www.lesswrong.com/posts/AcKRB8wDpdaN6v6ru/interpreting-gpt-the-logit-lens}, 2020.
\newblock [Accessed 09-02-2024].

\bibitem[Nye et~al.(2021)Nye, Andreassen, Gur-Ari, Michalewski, Austin, Bieber, Dohan, Lewkowycz, Bosma, Luan, et~al.]{nye2021scratchpad}
Maxwell Nye, Anders~Johan Andreassen, Guy Gur-Ari, Henryk Michalewski, Jacob Austin, David Bieber, David Dohan, Aitor Lewkowycz, Maarten Bosma, David Luan, et~al.
\newblock Show your work: Scratchpads for intermediate computation with language models.
\newblock \emph{arXiv preprint arXiv:2112.00114}, 2021.

\bibitem[Olsson et~al.(2022)Olsson, Elhage, Nanda, Joseph, DasSarma, Henighan, Mann, Askell, Bai, Chen, et~al.]{olsson2022icl-induction-heads}
Catherine Olsson, Nelson Elhage, Neel Nanda, Nicholas Joseph, Nova DasSarma, Tom Henighan, Ben Mann, Amanda Askell, Yuntao Bai, Anna Chen, et~al.
\newblock In-context learning and induction heads.
\newblock \emph{arXiv preprint arXiv:2209.11895}, 2022.

\bibitem[OpenAI(2024)]{openai2024gpt4}
OpenAI.
\newblock Gpt-4 technical report, 2024.

\bibitem[Prystawski et~al.(2024)Prystawski, Li, and Goodman]{prystawski2024think}
Ben Prystawski, Michael Li, and Noah Goodman.
\newblock Why think step by step? reasoning emerges from the locality of experience.
\newblock \emph{Advances in Neural Information Processing Systems}, 36, 2024.

\bibitem[Saparov \& He(2023)Saparov and He]{saparov2023prontoqa}
Abulhair Saparov and He~He.
\newblock Language models are greedy reasoners: A systematic formal analysis of chain-of-thought.
\newblock In \emph{The Eleventh International Conference on Learning Representations}, 2023.
\newblock URL \url{https://openreview.net/forum?id=qFVVBzXxR2V}.

\bibitem[Su et~al.(2023)Su, Lu, Pan, Murtadha, Wen, and Liu]{su2023roformer}
Jianlin Su, Yu~Lu, Shengfeng Pan, Ahmed Murtadha, Bo~Wen, and Yunfeng Liu.
\newblock Roformer: Enhanced transformer with rotary position embedding, 2023.

\bibitem[Tan(2023)]{tan2023causal-CoT}
Juanhe~TJ Tan.
\newblock Causal abstraction for chain-of-thought reasoning in arithmetic word problems.
\newblock In \emph{Proceedings of the 6th BlackboxNLP Workshop: Analyzing and Interpreting Neural Networks for NLP}, pp.\  155--168, 2023.

\bibitem[Touvron et~al.(2023)Touvron, Martin, Stone, Albert, Almahairi, Babaei, Bashlykov, Batra, Bhargava, Bhosale, Bikel, Blecher, Ferrer, Chen, Cucurull, Esiobu, Fernandes, Fu, Fu, Fuller, Gao, Goswami, Goyal, Hartshorn, Hosseini, Hou, Inan, Kardas, Kerkez, Khabsa, Kloumann, Korenev, Koura, Lachaux, Lavril, Lee, Liskovich, Lu, Mao, Martinet, Mihaylov, Mishra, Molybog, Nie, Poulton, Reizenstein, Rungta, Saladi, Schelten, Silva, Smith, Subramanian, Tan, Tang, Taylor, Williams, Kuan, Xu, Yan, Zarov, Zhang, Fan, Kambadur, Narang, Rodriguez, Stojnic, Edunov, and Scialom]{touvron2023llama}
Hugo Touvron, Louis Martin, Kevin Stone, Peter Albert, Amjad Almahairi, Yasmine Babaei, Nikolay Bashlykov, Soumya Batra, Prajjwal Bhargava, Shruti Bhosale, Dan Bikel, Lukas Blecher, Cristian~Canton Ferrer, Moya Chen, Guillem Cucurull, David Esiobu, Jude Fernandes, Jeremy Fu, Wenyin Fu, Brian Fuller, Cynthia Gao, Vedanuj Goswami, Naman Goyal, Anthony Hartshorn, Saghar Hosseini, Rui Hou, Hakan Inan, Marcin Kardas, Viktor Kerkez, Madian Khabsa, Isabel Kloumann, Artem Korenev, Punit~Singh Koura, Marie-Anne Lachaux, Thibaut Lavril, Jenya Lee, Diana Liskovich, Yinghai Lu, Yuning Mao, Xavier Martinet, Todor Mihaylov, Pushkar Mishra, Igor Molybog, Yixin Nie, Andrew Poulton, Jeremy Reizenstein, Rashi Rungta, Kalyan Saladi, Alan Schelten, Ruan Silva, Eric~Michael Smith, Ranjan Subramanian, Xiaoqing~Ellen Tan, Binh Tang, Ross Taylor, Adina Williams, Jian~Xiang Kuan, Puxin Xu, Zheng Yan, Iliyan Zarov, Yuchen Zhang, Angela Fan, Melanie Kambadur, Sharan Narang, Aurelien Rodriguez, Robert Stojnic, Sergey Edunov, and Thomas
  Scialom.
\newblock Llama 2: Open foundation and fine-tuned chat models, 2023.

\bibitem[Vaswani et~al.(2017)Vaswani, Shazeer, Parmar, Uszkoreit, Jones, Gomez, Kaiser, and Polosukhin]{transformer}
Ashish Vaswani, Noam Shazeer, Niki Parmar, Jakob Uszkoreit, Llion Jones, Aidan~N Gomez, \L~ukasz Kaiser, and Illia Polosukhin.
\newblock Attention is all you need.
\newblock In \emph{Advances in Neural Information Processing Systems}, volume~30, 2017.
\newblock URL \url{https://proceedings.neurips.cc/paper_files/paper/2017/file/3f5ee243547dee91fbd053c1c4a845aa-Paper.pdf}.

\bibitem[Wang et~al.(2023)Wang, Variengien, Conmy, Shlegeris, and Steinhardt]{wang2023interpretability}
Kevin~Ro Wang, Alexandre Variengien, Arthur Conmy, Buck Shlegeris, and Jacob Steinhardt.
\newblock Interpretability in the wild: a circuit for indirect object identification in {GPT}-2 small.
\newblock In \emph{The Eleventh International Conference on Learning Representations}, 2023.
\newblock URL \url{https://openreview.net/forum?id=NpsVSN6o4ul}.

\bibitem[Wang \& Wang(2023)Wang and Wang]{wang2023reasoning}
Xinyi Wang and William~Yang Wang.
\newblock Reasoning ability emerges in large language models as aggregation of reasoning paths: A case study with knowledge graphs.
\newblock In \emph{Workshop on Efficient Systems for Foundation Models@ ICML2023}, 2023.

\bibitem[Wei et~al.(2022{\natexlab{a}})Wei, Tay, Bommasani, Raffel, Zoph, Borgeaud, Yogatama, Bosma, Zhou, Metzler, Chi, Hashimoto, Vinyals, Liang, Dean, and Fedus]{wei2022emergent}
Jason Wei, Yi~Tay, Rishi Bommasani, Colin Raffel, Barret Zoph, Sebastian Borgeaud, Dani Yogatama, Maarten Bosma, Denny Zhou, Donald Metzler, Ed~H. Chi, Tatsunori Hashimoto, Oriol Vinyals, Percy Liang, Jeff Dean, and William Fedus.
\newblock Emergent abilities of large language models.
\newblock \emph{Transactions on Machine Learning Research}, 2022{\natexlab{a}}.
\newblock ISSN 2835-8856.
\newblock URL \url{https://openreview.net/forum?id=yzkSU5zdwD}.
\newblock Survey Certification.

\bibitem[Wei et~al.(2022{\natexlab{b}})Wei, Wang, Schuurmans, Bosma, ichter, Xia, Chi, Le, and Zhou]{CoT-prompting-2022}
Jason Wei, Xuezhi Wang, Dale Schuurmans, Maarten Bosma, brian ichter, Fei Xia, Ed~Chi, Quoc~V Le, and Denny Zhou.
\newblock Chain-of-thought prompting elicits reasoning in large language models.
\newblock In S.~Koyejo, S.~Mohamed, A.~Agarwal, D.~Belgrave, K.~Cho, and A.~Oh (eds.), \emph{Advances in Neural Information Processing Systems}, volume~35, pp.\  24824--24837. Curran Associates, Inc., 2022{\natexlab{b}}.
\newblock URL \url{https://proceedings.neurips.cc/paper_files/paper/2022/file/9d5609613524ecf4f15af0f7b31abca4-Paper-Conference.pdf}.

\bibitem[Wu et~al.(2023)Wu, Geiger, Potts, and Goodman]{wu2023interpretability}
Zhengxuan Wu, Atticus Geiger, Christopher Potts, and Noah~D Goodman.
\newblock Interpretability at scale: Identifying causal mechanisms in alpaca.
\newblock \emph{arXiv preprint arXiv:2305.08809}, 2023.

\bibitem[Zhang \& Nanda(2023)Zhang and Nanda]{zhang2023patching}
Fred Zhang and Neel Nanda.
\newblock Towards best practices of activation patching in language models: Metrics and methods.
\newblock \emph{arXiv preprint arXiv:2309.16042}, 2023.

\end{thebibliography}
\bibliographystyle{tmlr}
\newpage
\appendix
\section{PrOntoQA question example}
\label{app:subsec:prontoqa-example}
\begin{figure}[!t]
    \centering
    \includegraphics[width=0.7\textwidth]{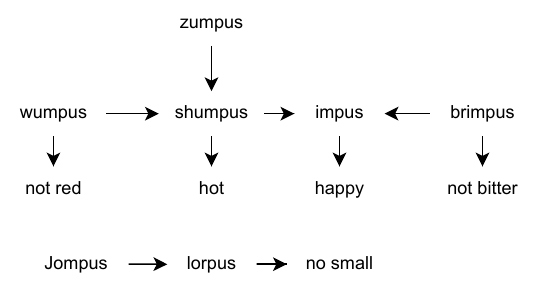}
    \caption{An example of fictional ontology in the PrOntoQA dataset.}
    \label{fig:prontoqa-tree}
\end{figure}

Following are two QA examples from fictional and false ontologies, respectively:

\textbf{Fictional ontology example:}\\
\textit{Context:} Tumpus is bright. Lempus is tumpus. Max is lempus.\\
\textit{Query:} True or false: Max is bright.\\
\textit{Answer:} Max is lempus. Lempus is tumpus. Max is tumpus. Tumpus is bright. Max is bright. True\\
\textbf{False onotology example:}\\
\textit{Context:} Carnivore is spicy. Vertebrates are carnivores. Rex is vertebrate.\\
\textit{Query:} True or false: Rex is spicy.\\
\textit{Answer:} Rex is vertebrate. Vertebrates are carnivores. Rex is carnivore. Carnivore is spicy. Rex is spicy. True

Figure~\ref{fig:prontoqa-tree} demonstrates an example ontology tree along with distractors. Here, {\em zumpus} \(\rightarrow\) {\em shumpus} translates to {\em zumpuses are shumpuses}. From this example, one can construct the following multistep reasoning problem: {\em Wumpuses are shumpuses. Shumpuses are impuses. Max is a wumpus. Wumpuses are not red. Impuses are happy. Jompuses are lorpuses. True or False: Max is happy}. In this example, the statement {\em Jompuses are lorpuses} serves as a distractor as neither of the entities are part of the ontology relevant to the question.

\section{Knockout}
\label{app:sec:knockout}

To perform knockout average activation from false ontology the PrOntoQA dataset is used. For each subtask, false ontology activation is stored and mean activation knockout is performed. We provide step-by-step examples of subtask-wise generation over false ontology as follows:

\textit{Context:} Carnivore is spicy. Vertebrates are carnivores. Rex is vertebrate.\\
\textit{Query:} True or false: Rex is spicy.\\
\textit{Answer}: "Rex is vertebrate. Vertebrates are carnivores. Rex is carnivore. Carnivore is spicy. Rex is spicy. True",\\
"prompt\_0": "Carnivore is spicy. Vertebrates are carnivores. Rex is vertebrate. True or false: Rex is spicy. Let us think step by step.",\\
"prompt\_1": "Carnivore is spicy. Vertebrates are carnivores. Rex is vertebrate. True or false: Rex is spicy. Let us think step by step. Rex is",\\
"prompt\_2": "Carnivore is spicy. Vertebrates are carnivores. Rex is vertebrate. True or false: Rex is spicy. Let us think step by step. Rex is vertebrate.",\\
"prompt\_3": "Carnivore is spicy. Vertebrates are carnivores. Rex is vertebrate. True or false: Rex is spicy. Let us think step by step. Rex is vertebrate. Vertebrates are",\\
"prompt\_4": "Carnivore is spicy. Vertebrates are carnivores. Rex is vertebrate. True or false: Rex is spicy. Let us think step by step. Rex is vertebrate. Vertebrates are carnivores.",\\
"prompt\_5": "Carnivore is spicy. Vertebrates are carnivores. Rex is vertebrate. True or false: Rex is spicy. Let us think step by step. Rex is vertebrate. Vertebrates are carnivores. Rex is",\\
"prompt\_6": "Carnivore is spicy. Vertebrates are carnivores. Rex is vertebrate. True or false: Rex is spicy. Let us think step by step. Rex is vertebrate. Vertebrates are carnivores. Rex is carnivore.",\\
"prompt\_7": "Carnivore is spicy. Vertebrates are carnivores. Rex is vertebrate. True or false: Rex is spicy. Let us think step by step. Rex is vertebrate. Vertebrates are carnivores. Rex is carnivore. Carnivore is",\\
"prompt\_8": "Carnivore is spicy. Vertebrates are carnivores. Rex is vertebrate. True or false: Rex is spicy. Let us think step by step. Rex is vertebrate. Vertebrates are carnivores. Rex is carnivore. Carnivore is spicy.",\\
"prompt\_9": "Carnivore is spicy. Vertebrates are carnivores. Rex is vertebrate. True or false: Rex is spicy. Let us think step by step. Rex is vertebrate. Vertebrates are carnivores. Rex is carnivore. Carnivore is spicy. Rex is spicy.",

\section{Task specific head identification}
\label{app:task-specific-heads}

\subsection{Decision-making heads}

To identify the heads that actively participate in decision-making subtasks, we incorporate activation patching on individual heads over decision-making subtasks.

Let \(S_\text{Decision}\) denote the input token sequence for a particular decision-making subtask with \(s_\text{ans}\) being the token corresponding to the correct answer. Also, let \(P_\text{org}\) denote the output token probability distribution of the full model corresponding to \(S_\text{Decision}\) as input, i.e.,
\begin{align*}
    P_\text{org} = \operatorname{SoftMax}\left (\operatorname{LM}(\vx_\text{logit}|S_\text{Decision})\right )
\end{align*}
Here we omit the token position superscript over \(\vx_\text{logit}\) for brevity since we are taking the output corresponding to the last input token. We store the activations \(\vy_{j,k}\) corresponding to each head \(h_{j,k}\).

Next, we corrupt the input corresponding to the last token in \(S_\text{Decision}\), which is equivalent to knocking off the set of all heads, \(\mathcal{H}_\text{full}\). We record the corresponding output token probability as,
\begin{align*}
    P_\text{corrupt} = \operatorname{SoftMax}\left ( \operatorname{LM}_{\mathcal{H}_\text{full}} \left (\vx_\text{logit}|S_\text{Decision}\right) \right)
\end{align*}

Finally, for each head \(h_{j,k}\), we restore its corresponding original output \(\vy_{j,k}\) and record the output token probability distribution as
\begin{align*}
    P^{j,k}_\text{patched} = \operatorname{SoftMax} \left (\operatorname{LM}_{\mathcal{H}_\text{full}\setminus\{h_{j,k}\}}\left (\vx_\text{logit}|S_\text{Decision}\right) \right)
\end{align*}

Then, we compute the importance score of head \(h_{j,k}\) as
\begin{align*}
\mu_\text{Decision}(h_{j,k}) = \frac{P_\text{org}(x=s_\text{ans}) - P_\text{corrupt}(x=s_\text{ans})}{P_\text{org}(x=s_\text{ans}) - P_\text{patched}(x=s_\text{ans})}    
\end{align*}

\subsection{Copy heads}

Copying subtasks follows decision-making subtasks immediately. Once the head entity of a statement is decided, copying requires moving the tail entity corresponding to that statement in the input context to the output (e.g., if there is a statement {\em Rompus is grimpus} in the input context and decision-making heads have decided to output {\em Rompus}, then copy subtask requires moving {\em grimpus} to the output). For this, we simply look into the attention probability assigned by each head \(h_{j,k}\) to the token to be copied as source (i.e., key), denoted as
\begin{align*}
    \mu_\text{Copy}(h_{j,k}) = \frac{\exp\left(  (\mW^{j,k}_Q\vx^\text{end}_j)^\top (\mW^{j,k}_K\vx^\text{ans}_j) \right)}{\sum_i \exp\left(  (\mW^{j,k}_Q\vx^\text{end}_j)^\top (\mW^{j,k}_K\vx^i_j) \right)}
\end{align*}
where \(\vx^\text{end}_j, \vx^\text{ans}_j\) denote the residual streams corresponding to the last token and the answer token, respectively, at \(j\)-th decoder block.

\subsection{Inductive reasoning heads}

Let the input token sequence for the inductive reasoning subtask be denoted as \(S_\text{ind}\), which is of the form {\tt [A] is [B]. [B] is [C]. [A] is}, and let \(s_\text{ans}\) denote the answer token (which is {\tt [C]} in this example). We represent the first and second occurrences of each token {\tt [A]} and {\tt [B]} as \(A_1, A_2\) and \(B_1, B_2\), respectively. For a given \(S_\text{Induction}\), we can perform activation patching on three different residual streams, \(\vx^{B_1}_j\), \(\vx^{B_2}_j\), and \(\vx^{C}_j\) to identify the responsible heads. Let the original token probability distribution be 
\begin{align*}
    P_\text{org} = \operatorname{SoftMax}\left (\operatorname{LM}(\vx_\text{logit}|S_\text{Induction})\right )
\end{align*}

Next, for each \(l\in \{B_1, B_2, C\}\), we compute the corrupted forward pass as follows:
\begin{align*}
    P_{\text{corrupt},l} = \operatorname{SoftMax}\left (\operatorname{LM}^l_{\mathcal{H}_\text{full}}(\vx_\text{logit}|S_\text{Induction})\right )
\end{align*}
followed by patching original activation to the head \(h_{j,k}\) as 
\begin{align*}
    P^{j,k}_{\text{patched},l} = \operatorname{SoftMax}\left (\operatorname{LM}^l_{\mathcal{H}_\text{full}\setminus\{h_{j,k}\}}(\vx_\text{logit}|S_\text{Induction})\right )
\end{align*}
Then, the inductive reasoning head importance is computed as:
\begin{align*}
    \mu_\text{Induction}(h_{j,k}) = \operatorname{Mean}_l\frac{D^l_\text{ref}-D^l_\text{patched}}{D^l_\text{ref}}
\end{align*}
where \(D^l_\text{ref}\) is the KL divergence between \(P_\text{org}\) and \(P_{\text{corrupt},l}\), and \(D^l_\text{patched}\) is the KL divergence between \(P_\text{org}\) and \(P_{\text{corrupt},l}\).

\subsection{Performance of inductive reasoning heads for each subtask}

For each subtask 0-9, we group the attention heads based on their respective \(\mu_\text{Induction}(h_{j,k})\). We knockout heads within a certain threshold head importance range \((\mu_\text{min}, \mu_\text{max})\) by knockout and record the accuracy. In Table~\ref{tab:head-removal}, we report the attention head statistics that we used in the analysis for context abidance and answer-writing pathways. Note that we sought to keep the relative accuracy (i.e., the fraction of correct prediction by the ablated model over the correct predictions of the full model) close to \(0.9\).
 
\begin{table}[!ht]
    \centering
    \begin{tabular}{|l|l|l|l|l|}
    \hline
        Subtask index & Accuracy & Heads removed & Threshold range \\ \hline
        0 & 1 & 475 & 0.30487806 - 0.31463414 \\ \hline
        1 & 0.93 & 554 & 0.30243903,0.31707317\\ \hline
        2 & 0.96 & 617 & 0.3 - 0.3195122\\ \hline
        3 & 1 & 617 & 0.3 - 0.3195122\\ \hline
        4 & 0.99 & 554 & 0.30243903 - 0.31707317\\ \hline
        5 & 0.93 & 475 & 0.30487806 - 0.31463414\\ \hline
        6 & 0.94 & 475 & 0.30487806 - 0.31463414\\ \hline
        7 & 0.88 & 617 & 0.3 - 0.3195122\\ \hline
        8 & 0.95 & 617 & 0.3 - 0.3195122\\ \hline
        9 & 0.91 & 663 & 0.297561 - 0.3219512 \\ \hline
    \end{tabular}
    \caption{Statistics of subtask-wise attention head removal according to their importance in inductive reasoning.}
    \label{tab:head-removal}
\end{table}

\section{Probing for token mixing}
\label{app:sec:probing}

Towards probing ontological relatedness in residual streams (see Section~\ref{sec:token-mixing}), first, we need to extract the positive, negative and unrelated entities from the sentence. PrOntoQA provides us with a tree structure for each sentence, and with the help of the structure, we can extract the pairs. Consider the following example.

Sentence: {\em Each shumpus is a zumpus. Shumpuses are wumpuses. Every shumpus is hot. Impuses are shumpuses. Each impus is a brimpus. Each impus is happy. Wumpuses are not red. Brimpuses are not bitter. Lorpuses are jompuses. Each yumpus is not hot. Lorpuses are not small. Max is an impus. Max is a lorpus.}

Here all immediate entities will be termed as posive pairs, such as {\em shumpus} <> {\em impus}, {\em zumpus} <> {\em shumpus}, {\em impus}<> {\em happy}, etc. Similarly, following are a few examples of negatively related entities: {\em wumpus} <> {\em red}, {\em brimpus <> bitter}, etc. The distractor entities (jompus, lorpus and small) serves as seed for unrelated entities; any two entities that belong two disjoint ontologies are taken as unrelated.

During training and testing, we also make sure that there are no overlap between entities. During training, we use entities such as "wumpus", "yumpus", "zumpus", "dumpus" and during testing : "zonkify", "quiblitz","flimjam", "zizzlewump", "snickerblat".
Following are the implementation details for the probing classifier:

\begin{itemize}
    \item Total training pairs: 28392; Total testing pairs: 9204. All three types of pairs (positively and negatively related and unrelated) are present in equal proportion in the training and testing data.
    \item 4-layer MLP model, 4096 * 2 -> 128 -> 64 -> 32 -> 3. With ReLU in between each Linear layer.
    \item Learning rate: 0.00005
    \item Number of epochs: 120.
\end{itemize}

\section{Prompts used and model performance}
\label{app:sec:prompts-and-model}

We use 6-shot examples of CoT for generation in all the experiments.:

\#\#\# Input:\\
Gorpus is twimpus. Alex is rompus. Rompus is gorpus. Gorpus is small. Rompus is mean. True or false: Alex is small. Let us think step by step.\\
\#\#\# Response:\\
Alex is rompus. Rompus is gorpus. Alex is gorpus. Gorpus is small. Alex is small. True

\#\#\# Input:\\
Gorpuses are discordant. Max is zumpus. Zumpus is shampor. Zumpus is gorpus. Gorpus is earthy. True or false: Max is small. Let us think step by step. \\
\#\#\# Response:\\
Max is zumpus. Zumpus is gorpus. Max is gorpus. Gorpuses are discordant. Max is discordant. False

\#\#\# Input:\\
Borpin are wumpus. Wumpuses are angry. Wumpus is jempor. Sally is lempus. Lempus is wumpus. True or false: Sally is floral. Let us think step by step. \\
\#\#\# Response:\\
Sally is lempus. Lempus is wumpus. Sally is wumpus. Wumpuses are angry. Sally is angry. False

\#\#\# Input:\\
Gorpus is jelgit. Yumpuses are loud. Gorpus is yumpus. Yumpus is orange. Rex is gorpus. True or false: Rex is loud. Let us think step by step.\\
\#\#\# Response:\\
Rex is gorpus. Gorpus is yumpus. Rex is yumpus. Yumpuses are loud. Rex is loud. True

\#\#\# Input:\\
Lempus is tumpus. Max is lempus. Tumpus is fruity. Tumpus is bright. Lempus is dropant. True or false: Max is bright. Let us think step by step.\\
\#\#\# Response:\\
Max is lempus. Lempus is tumpus. Max is tumpus. Tumpus is bright. Max is bright. True

\#\#\# Input:\\
Sterpuses are dull. Impus is medium. Impuses are sterpuses. Wren is impus. Sterpus is daumpin. True or false: Wren is melodic. Let us think step by step.\\
\#\#\# Response:\\
Wren is impus. Impuses are sterpuses. Wren is sterpus. Sterpuses are dull. Wren is dull. False

To ensure that the model does not learn any positional pattern from the few-shot example, we jumble the statements in the Input context and add two extra distractors in the statement.

Following are the performance of {\color{black} Llama-2}-7B on two subdomains of PrOntoQA ---
(i) accuracy on false ontology: 52.5, (ii) accuracy on fictional ontology: 35.9.

Accuracy is calculated by a complete match of all the intermediate reasoning subtasks and the final True/False statement. We make the following
modification to the PrOntoQA dataset:
(i) removal of articles like {\em a} and {\em an} from statements, (ii) removal of {\em every} and {\em each} from statements, (iii) replacing negative entities with incorrect entities (e.g., {\em Max is not cold} \(\rightarrow\) {\em Max is dog}).

\section{Attention probability}
\label{app:sec:attention}
\begin{figure}[!t]
    \centering
    \includegraphics[width=\textwidth]{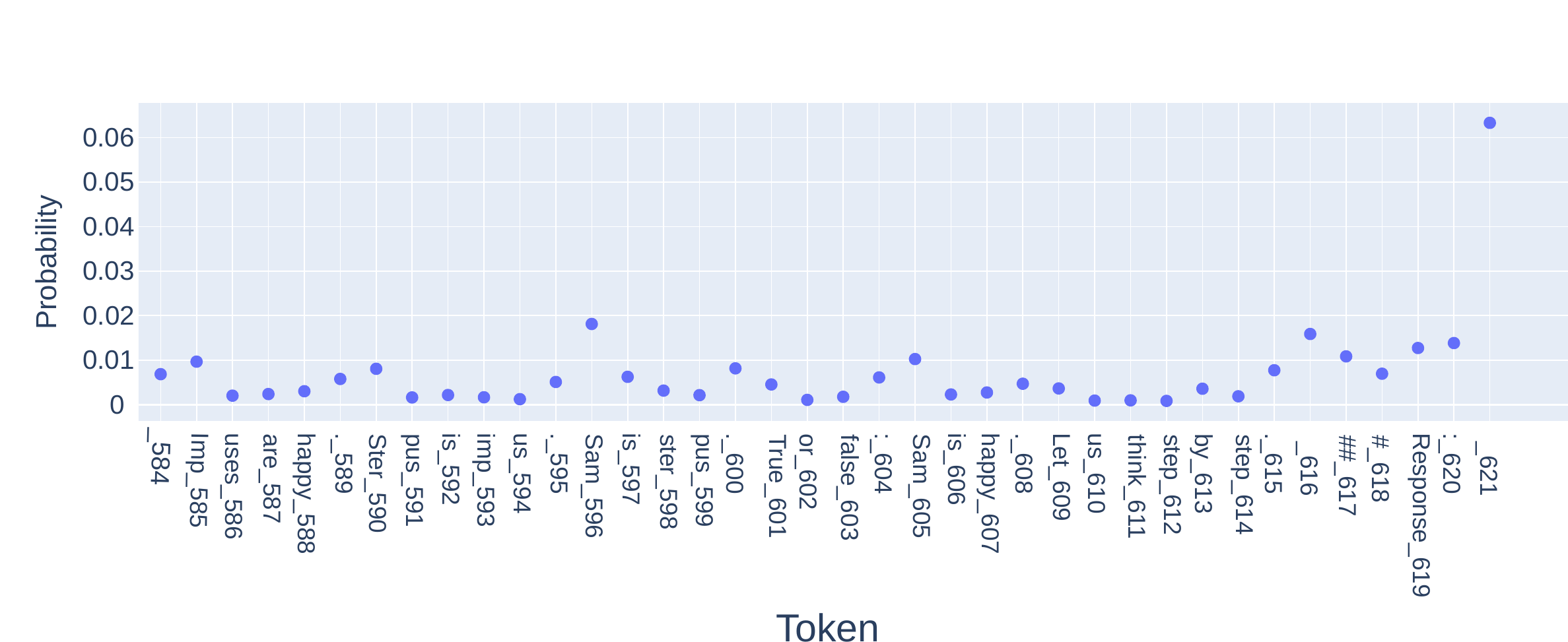}
    \caption{Average attention probability distribution (across all heads) over question context in an example subtask 0. Here, the model needs to predict the token {\em Sam}.}
    \label{fig:attn-Input}
\end{figure}

\begin{figure}
    \centering
    \begin{minipage}[t]{0.76\textwidth}
    \vspace{0pt}\includegraphics[width=\textwidth]{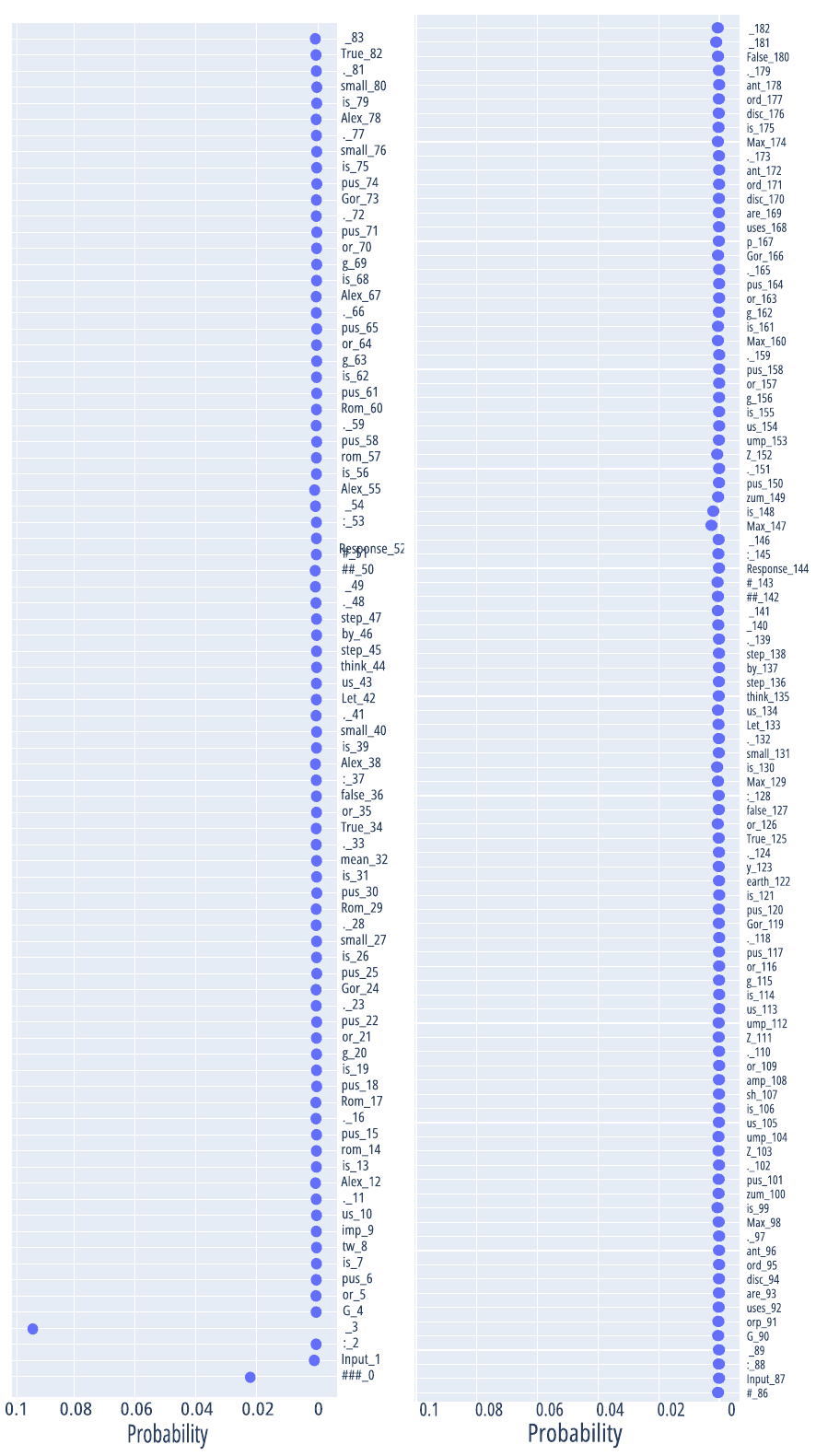}
    \end{minipage}\hfill
    \begin{minipage}[t]{0.23\textwidth}
    \vspace{0pt}
    \caption{{\bf Attention probability distribution over few-shot examples}. In this plot, we show the average attention scores over tokens corresponding to the first (left) and second (right) examples (for the rest of the examples, see Figure~\ref{fig:few-shot-att-3-4} and \ref{fig:few-shot-att-5-6}) among the 6-shot context corresponding to the question in Figure~\ref{fig:attn-Input}. Here, the model is at subtask-0 and needs to predict {\em Sam}. A slight peak in attention probability at the token corresponding to subtask-0 in the second example ({\em Max} in this case) is observable.}
    \label{fig:few-shot-att-1-2}
    \end{minipage}
\end{figure}

\begin{figure}
    \centering
    \begin{minipage}[t]{0.76\textwidth}
    \vspace{0pt}\includegraphics[width=\textwidth]{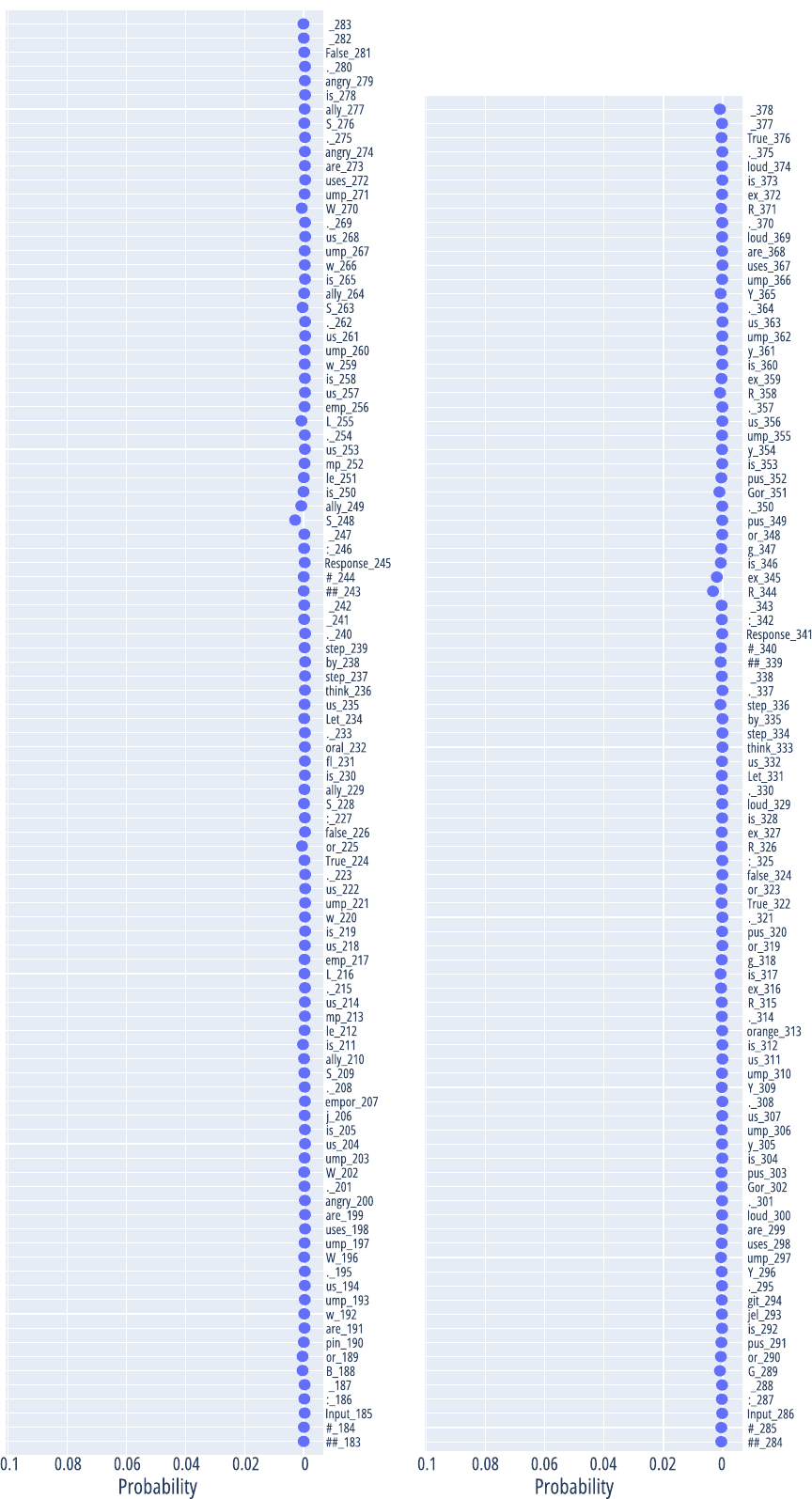}
    \end{minipage}\hfill
    \begin{minipage}[t]{0.23\textwidth}
    \vspace{0pt}
    \caption{{\bf Attention probability distribution over few-shot examples} (continued after Figure~\ref{fig:few-shot-att-1-2}). In this plot, we show the average attention scores over tokens corresponding to the third (left) and fourth (right) examples among the few-shot context corresponding to the question in Figure~\ref{fig:attn-Input}. Here, the model is at subtask-0 and needs to predict {\em Sam}. Small peaks in attention probability at tokens corresponding to subtask-0 in the third ({\em S} from {\em Sally}) and fourth ({\em R} from {\em Rex}) example are observable. For the rest of the examples, see Figure~\ref{fig:few-shot-att-5-6}}
    \label{fig:few-shot-att-3-4}
    \end{minipage}
\end{figure}

\begin{figure}
    \centering
    \begin{minipage}[t]{0.76\textwidth}
    \vspace{0pt}\includegraphics[width=\textwidth]{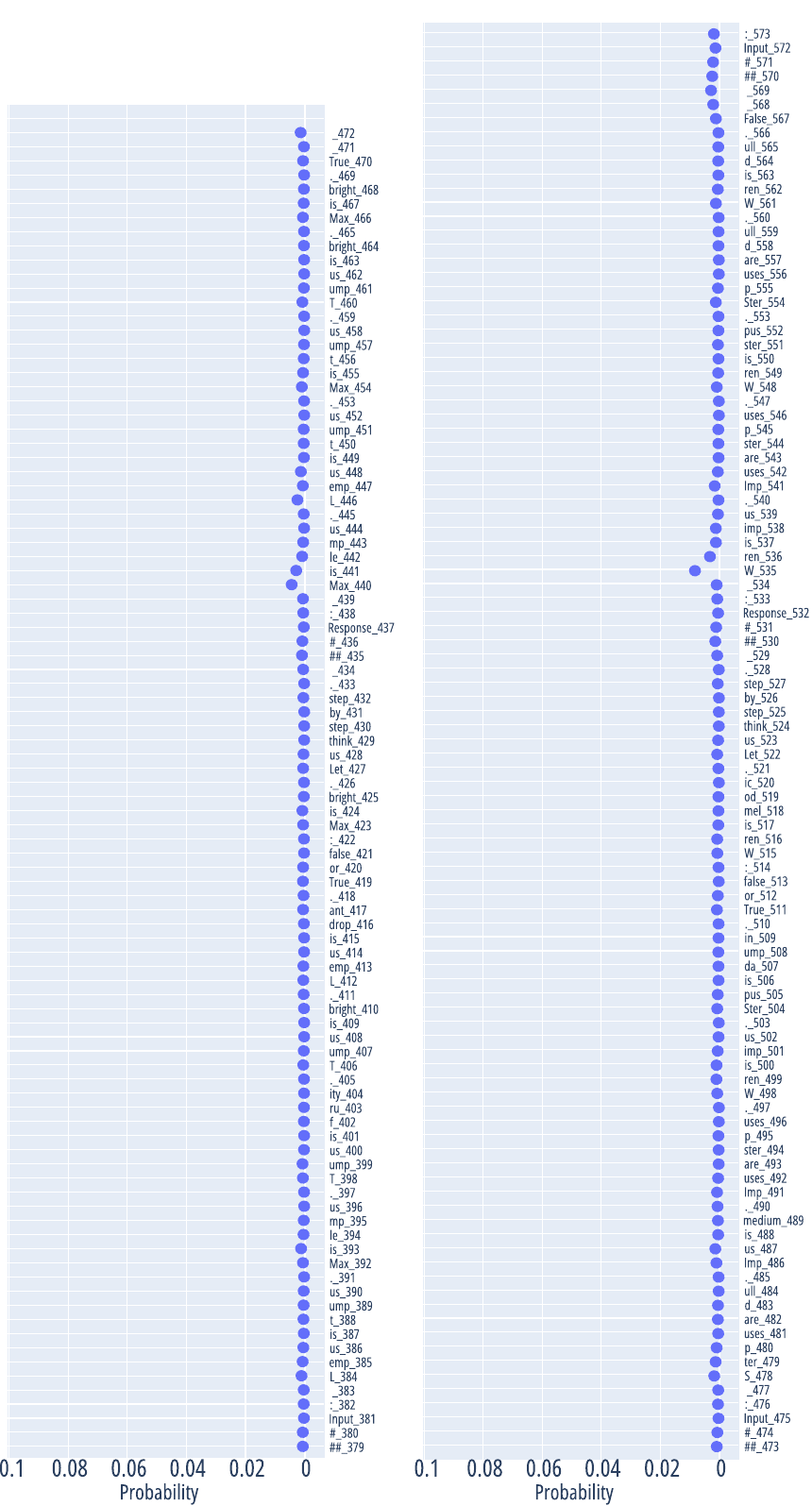}
    \end{minipage}\hfill
    \begin{minipage}[t]{0.23\textwidth}
    \vspace{0pt}
    \caption{{\bf Attention probability distribution over few-shot examples} (continued after Figure~\ref{fig:few-shot-att-1-2} and \ref{fig:few-shot-att-3-4}). In this plot, we show the average attention scores over tokens corresponding to the fifth (left) and sixth (right) examples among the few-shot context corresponding to the question in Figure~\ref{fig:attn-Input}. Here, the model is at subtask-0 and needs to predict {\em Sam}. Small peaks in attention probability at tokens corresponding to subtask-0 in the fifth ({\em Max}) and sixth ({\em W} from {\em Wren}) example are observable.}
    \label{fig:few-shot-att-5-6}
    \end{minipage}
\end{figure}

We take the average of attention probability at the last residual stream of all the heads. Figure \ref{fig:attn-Input} shows the attention probability over the input context tokens, and Figures \ref{fig:few-shot-att-1-2}, \ref{fig:few-shot-att-3-4}, and \ref{fig:few-shot-att-5-6} show attention probability over the 6-shot examples used. This example is for subtask-0, and it can be seen that there are spikes in attention probability score for tokens at the same position as subtask 0, i.e., token just after "Response:". Similar patterns are observed for all the subtasks. Furthermore, the absolute value of attention to the task-specific tokens in the few-shot context increases as we move from the 1st to 6th example. 


\section{Information flow}
\label{app:sec:info-flow}

Assume a given head \(h_{j,k}\) that writes some information to the residual stream corresponding to the token \(s_i\) via output \(\vy^i_{j,k}\). Furthermore, let the maximum attention weight allocated by \(h_{j,k}\) be placed on the residual stream corresponding to token \(s_l\). Then, we represent the head with the triplet \((s_l, s_i, s_\text{info}=\arg\max \mU\vy^i_{j,k})\) denoting the source residual stream (source of information written by the head), target residual stream (target of information written by the head), and the information that is being written. 

\begin{figure}[!t]
    \centering
    \includegraphics[width=\textwidth]{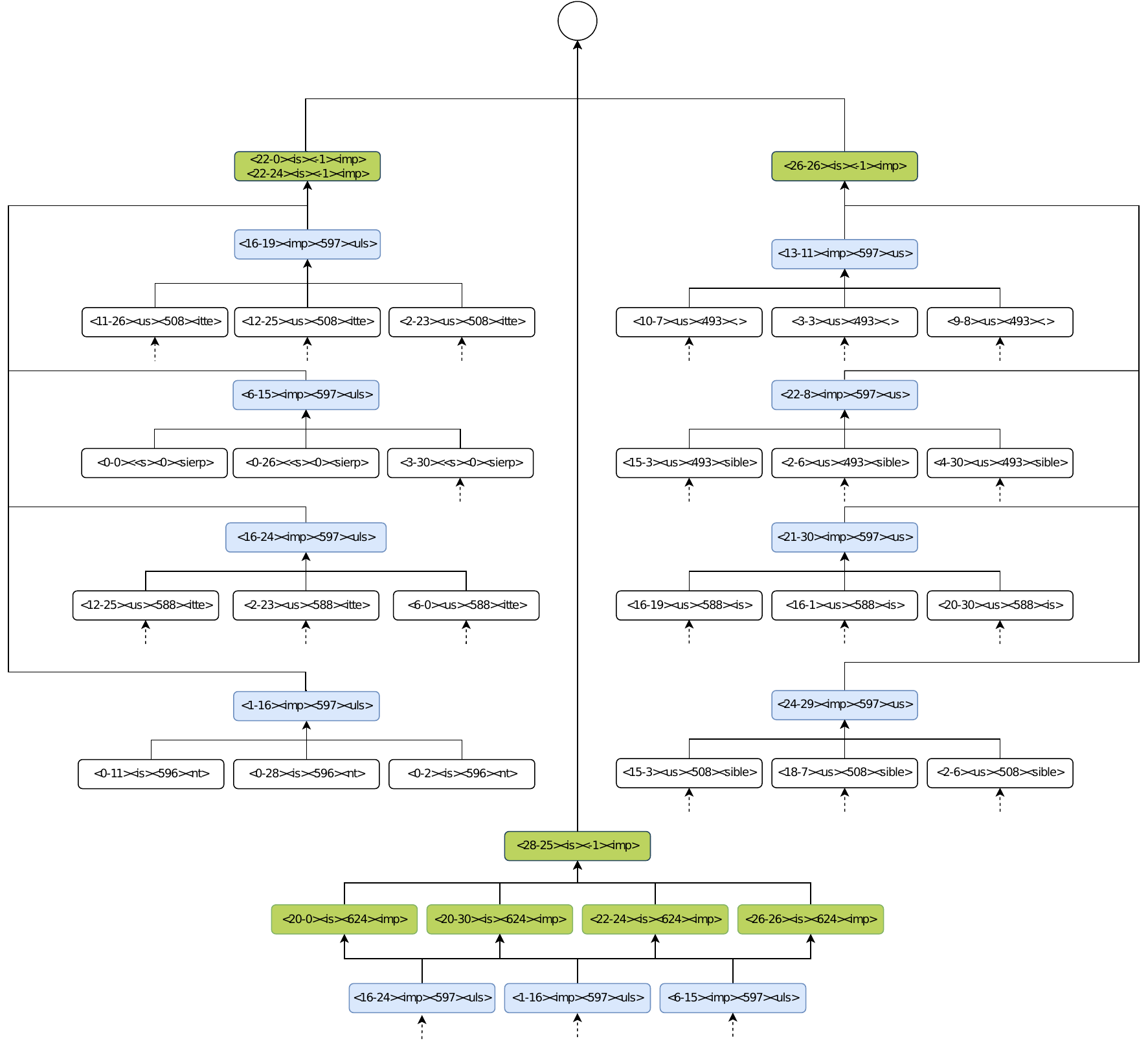}
    \caption{{\bf Infomation flow through attention heads towards answer-writing heads in {\color{black} Llama-2} 7B for subtask 1.} Each node in this DAG (pruned for brevity) is labeled as {\tt <layer index-head index><source residual content><source residual position><head output content}>. The subtask here is to predict {\em impus} given the generated response {\em Stella is} (see Appendix~\ref{app:sec:info-flow} for the complete tokenized input). In this example, there are seven answer writing heads (in {\color{green} green}) that are writing {\em <imp>} to the last residual stream corresponding to {\em is}. 
    In this particular example, the answer token {\em imp} is collected from  token no. 597 in the input question context by 8 different (in {\color{blue} blue}).}
    
    \label{fig:InformationFlow}
\end{figure}

We start with the answer writing heads, \(h^\text{Ans}_{j,k}\). For \((s_l, s_i, s_\text{info})\) associated with each \(h^\text{Ans}_{j,k}\), we identify the information content present within the source residual stream as \(\arg\max \mU\vx^l_{j-1}\) (i.e., the information present within the source residual stream right before the answer writer attended to it). Then, we look for all the heads in decoder blocks \(<j>\) responsible for writing that information into that residual stream. This continues till we reach a head that is in the starting decoder block. Figures~\ref{fig:InformationFlow} depict information flow extracted for a single example question for subtask 1. The tokenized input is as follows:

{\tt
<s>\_0
<s>\_1
<s>\_2
<s>\_3
<s>\_4
<s>\_5
\#\#\#\_6
Input\_7
:\_8
<0x0A>\_9
G\_10
or\_11
pus\_12
is\_13
tw\_14
imp\_15
us\_16
.\_17
Alex\_18
is\_19
rom\_20
pus\_21
.\_22
Rom\_23
pus\_24
is\_25
g\_26
or\_27
pus\_28
.\_29
Gor\_30
pus\_31
is\_32
small\_33
.\_34
Rom\_35
pus\_36
is\_37
mean\_38
.\_39
True\_40
or\_41
false\_42
:\_43
Alex\_44
is\_45
small\_46
.\_47
Let\_48
us\_49
think\_50
step\_51
by\_52
step\_53
.\_54
<0x0A>\_55
\#\#\_56
\#\_57
Response\_58
:\_59
<0x0A>\_60
Alex\_61
is\_62
rom\_63
pus\_64
.\_65
Rom\_66
pus\_67
is\_68
g\_69
or\_70
pus\_71
.\_72
Alex\_73
is\_74
g\_75
or\_76
pus\_77
.\_78
Gor\_79
pus\_80
is\_81
small\_82
.\_83
Alex\_84
is\_85
small\_86
.\_87
True\_88
<0x0A>\_89
<0x0A>\_90
\#\#\_91
\#\_92
Input\_93
:\_94
<0x0A>\_95
G\_96
orp\_97
uses\_98
are\_99
disc\_100
ord\_101
ant\_102
.\_103
Max\_104
is\_105
zum\_106
pus\_107
.\_108
Z\_109
ump\_110
us\_111
is\_112
sh\_113
amp\_114
or\_115
.\_116
Z\_117
ump\_118
us\_119
is\_120
g\_121
or\_122
pus\_123
.\_124
Gor\_125
pus\_126
is\_127
earth\_128
y\_129
.\_130
True\_131
or\_132
false\_133
:\_134
Max\_135
is\_136
small\_137
.\_138
Let\_139
us\_140
think\_141
step\_142
by\_143
step\_144
.\_145
\_146
<0x0A>\_147
\#\#\_148
\#\_149
Response\_150
:\_151
<0x0A>\_152
Max\_153
is\_154
zum\_155
pus\_156
.\_157
Z\_158
ump\_159
us\_160
is\_161
g\_162
or\_163
pus\_164
.\_165
Max\_166
is\_167
g\_168
or\_169
pus\_170
.\_171
Gor\_172
p\_173
uses\_174
are\_175
disc\_176
ord\_177
ant\_178
.\_179
Max\_180
is\_181
disc\_182
ord\_183
ant\_184
.\_185
False\_186
<0x0A>\_187
<0x0A>\_188
\#\#\_189
\#\_190
Input\_191
:\_192
<0x0A>\_193
B\_194
or\_195
pin\_196
are\_197
w\_198
ump\_199
us\_200
.\_201
W\_202
ump\_203
uses\_204
are\_205
angry\_206
.\_207
W\_208
ump\_209
us\_210
is\_211
j\_212
empor\_213
.\_214
S\_215
ally\_216
is\_217
le\_218
mp\_219
us\_220
.\_221
L\_222
emp\_223
us\_224
is\_225
w\_226
ump\_227
us\_228
.\_229
True\_230
or\_231
false\_232
:\_233
S\_234
ally\_235
is\_236
fl\_237
oral\_238
.\_239
Let\_240
us\_241
think\_242
step\_243
by\_244
step\_245
.\_246
\_247
<0x0A>\_248
\#\#\_249
\#\_250
Response\_251
:\_252
<0x0A>\_253
S\_254
ally\_255
is\_256
le\_257
mp\_258
us\_259
.\_260
L\_261
emp\_262
us\_263
is\_264
w\_265
ump\_266
us\_267
.\_268
S\_269
ally\_270
is\_271
w\_272
ump\_273
us\_274
.\_275
W\_276
ump\_277
uses\_278
are\_279
angry\_280
.\_281
S\_282
ally\_283
is\_284
angry\_285
.\_286
False\_287
<0x0A>\_288
<0x0A>\_289
\#\#\_290
\#\_291
Input\_292
:\_293
<0x0A>\_294
G\_295
or\_296
pus\_297
is\_298
jel\_299
git\_300
.\_301
Y\_302
ump\_303
uses\_304
are\_305
loud\_306
.\_307
Gor\_308
pus\_309
is\_310
y\_311
ump\_312
us\_313
.\_314
Y\_315
ump\_316
us\_317
is\_318
orange\_319
.\_320
R\_321
ex\_322
is\_323
g\_324
or\_325
pus\_326
.\_327
True\_328
or\_329
false\_330
:\_331
R\_332
ex\_333
is\_334
loud\_335
.\_336
Let\_337
us\_338
think\_339
step\_340
by\_341
step\_342
.\_343
<0x0A>\_344
\#\#\_345
\#\_346
Response\_347
:\_348
<0x0A>\_349
R\_350
ex\_351
is\_352
g\_353
or\_354
pus\_355
.\_356
Gor\_357
pus\_358
is\_359
y\_360
ump\_361
us\_362
.\_363
R\_364
ex\_365
is\_366
y\_367
ump\_368
us\_369
.\_370
Y\_371
ump\_372
uses\_373
are\_374
loud\_375
.\_376
R\_377
ex\_378
is\_379
loud\_380
.\_381
True\_382
<0x0A>\_383
<0x0A>\_384
\#\#\_385
\#\_386
Input\_387
:\_388
<0x0A>\_389
L\_390
emp\_391
us\_392
is\_393
t\_394
ump\_395
us\_396
.\_397
Max\_398
is\_399
le\_400
mp\_401
us\_402
.\_403
T\_404
ump\_405
us\_406
is\_407
f\_408
ru\_409
ity\_410
.\_411
T\_412
ump\_413
us\_414
is\_415
bright\_416
.\_417
L\_418
emp\_419
us\_420
is\_421
drop\_422
ant\_423
.\_424
True\_425
or\_426
false\_427
:\_428
Max\_429
is\_430
bright\_431
.\_432
Let\_433
us\_434
think\_435
step\_436
by\_437
step\_438
.\_439
<0x0A>\_440
\#\#\_441
\#\_442
Response\_443
:\_444
<0x0A>\_445
Max\_446
is\_447
le\_448
mp\_449
us\_450
.\_451
L\_452
emp\_453
us\_454
is\_455
t\_456
ump\_457
us\_458
.\_459
Max\_460
is\_461
t\_462
ump\_463
us\_464
.\_465
T\_466
ump\_467
us\_468
is\_469
bright\_470
.\_471
Max\_472
is\_473
bright\_474
.\_475
True\_476
<0x0A>\_477
<0x0A>\_478
\#\#\_479
\#\_480
Input\_481
:\_482
<0x0A>\_483
S\_484
ter\_485
p\_486
uses\_487
are\_488
d\_489
ull\_490
.\_491
Imp\_492
us\_493
is\_494
medium\_495
.\_496
Imp\_497
uses\_498
are\_499
ster\_500
p\_501
uses\_502
.\_503
W\_504
ren\_505
is\_506
imp\_507
us\_508
.\_509
Ster\_510
pus\_511
is\_512
da\_513
ump\_514
in\_515
.\_516
True\_517
or\_518
false\_519
:\_520
W\_521
ren\_522
is\_523
mel\_524
od\_525
ic\_526
.\_527
Let\_528
us\_529
think\_530
step\_531
by\_532
step\_533
.\_534
<0x0A>\_535
\#\#\_536
\#\_537
Response\_538
:\_539
<0x0A>\_540
W\_541
ren\_542
is\_543
imp\_544
us\_545
.\_546
Imp\_547
uses\_548
are\_549
ster\_550
p\_551
uses\_552
.\_553
W\_554
ren\_555
is\_556
ster\_557
pus\_558
.\_559
Ster\_560
p\_561
uses\_562
are\_563
d\_564
ull\_565
.\_566
W\_567
ren\_568
is\_569
d\_570
ull\_571
.\_572
False\_573
<0x0A>\_574
<0x0A>\_575
\#\#\_576
\#\_577
Input\_578
:\_579
<0x0A>\_580
J\_581
om\_582
pus\_583
is\_584
wooden\_585
.\_586
Imp\_587
us\_588
is\_589
j\_590
om\_591
pus\_592
.\_593
St\_594
ella\_595
is\_596
imp\_597
us\_598
.\_599
True\_600
or\_601
false\_602
:\_603
St\_604
ella\_605
is\_606
wooden\_607
.\_608
Let\_609
us\_610
think\_611
step\_612
by\_613
step\_614
.\_615
<0x0A>\_616
\#\#\_617
\#\_618
Response\_619
:\_620
<0x0A>\_621
St\_622
ella\_623
is\_624
}

\begin{figure}[!t]
    \centering
    \includegraphics[width=\textwidth]{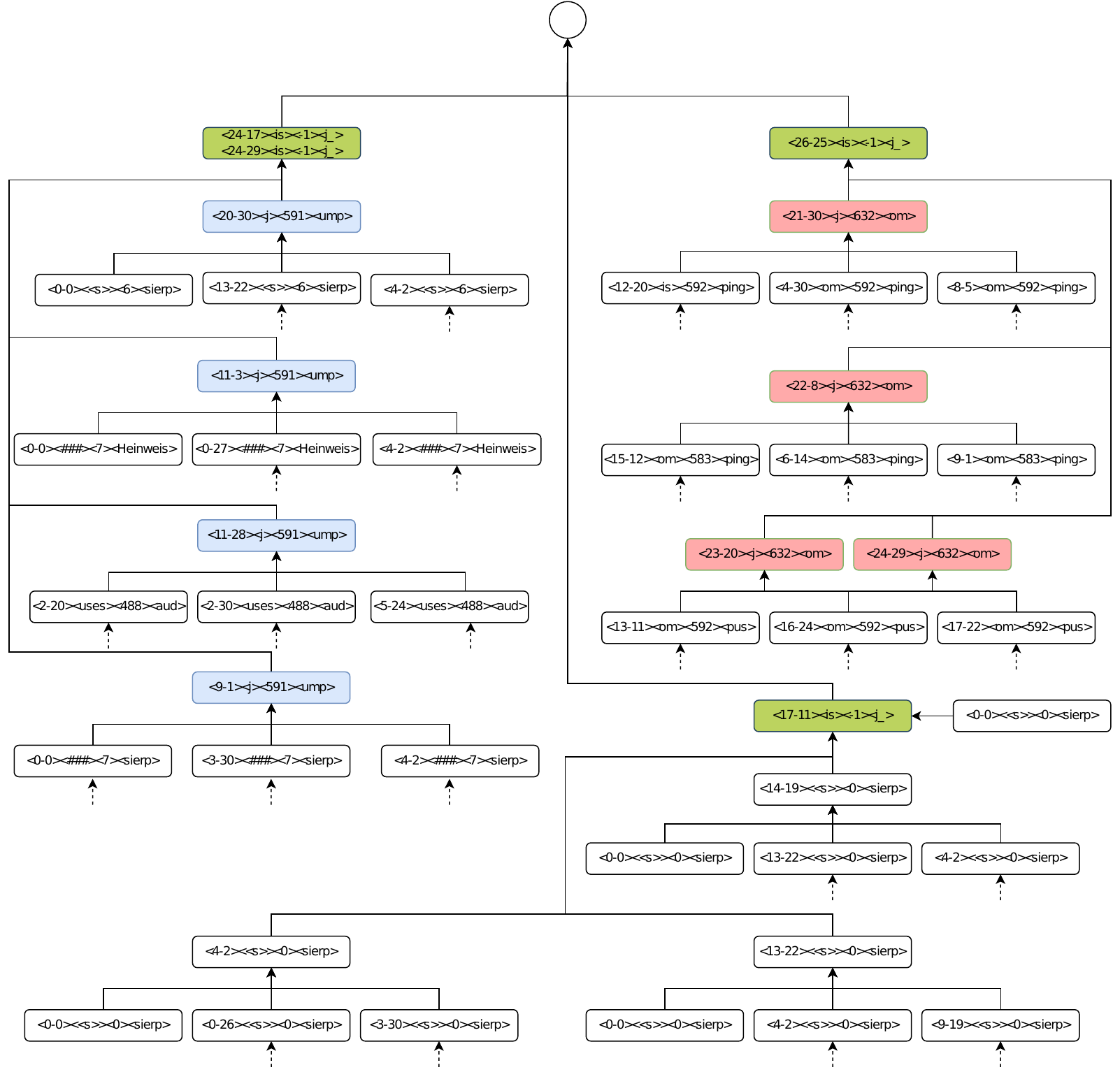}
    \caption{{\bf Infomation flow through attention heads towards answer-writing heads in {\color{black} Llama-2} 7B for subtask 5}. The representations of each node are similar to Figure \ref{fig:InformationFlow}. The subtask here is to predict {\em jompus} given the generated response {\em Stella is impus. Impus is jompus. Stella is} (see Appendix~\ref{app:sec:info-flow} for the complete tokenized input). In this example, there are three answer writing heads (in {\color{green} green}) that are writing {\em <j>} to the last residual stream corresponding to {\em is}. In this particular example, the answer token {\em imp} is collected from two position, token no. 591 in the input question context 4 heads (in {\color{blue} blue}), and token no. 632 in the generated context by 4 heads (in {\color{red} red}).
    }
    \label{fig:InformationFlow_noise_index_5}
\end{figure}

Figure ~\ref{fig:InformationFlow_noise_index_5} depict information flow extracted for a single example question for subtask 5. The tokenized input is as follows:

{\tt
<s>\_0
<s>\_1
<s>\_2
<s>\_3
<s>\_4
<s>\_5
<s>\_6
\#\#\#\_7
Input\_8
:\_9
<0x0A>\_10
G\_11
or\_12
pus\_13
is\_14
tw\_15
imp\_16
us\_17
.\_18
Alex\_19
is\_20
rom\_21
pus\_22
.\_23
Rom\_24
pus\_25
is\_26
g\_27
or\_28
pus\_29
.\_30
Gor\_31
pus\_32
is\_33
small\_34
.\_35
Rom\_36
pus\_37
is\_38
mean\_39
.\_40
True\_41
or\_42
false\_43
:\_44
Alex\_45
is\_46
small\_47
.\_48
Let\_49
us\_50
think\_51
step\_52
by\_53
step\_54
.\_55
<0x0A>\_56
\#\#\_57
\#\_58
Response\_59
:\_60
<0x0A>\_61
Alex\_62
is\_63
rom\_64
pus\_65
.\_66
Rom\_67
pus\_68
is\_69
g\_70
or\_71
pus\_72
.\_73
Alex\_74
is\_75
g\_76
or\_77
pus\_78
.\_79
Gor\_80
pus\_81
is\_82
small\_83
.\_84
Alex\_85
is\_86
small\_87
.\_88
True\_89
<0x0A>\_90
<0x0A>\_91
\#\#\_92
\#\_93
Input\_94
:\_95
<0x0A>\_96
G\_97
orp\_98
uses\_99
are\_100
disc\_101
ord\_102
ant\_103
.\_104
Max\_105
is\_106
zum\_107
pus\_108
.\_109
Z\_110
ump\_111
us\_112
is\_113
sh\_114
amp\_115
or\_116
.\_117
Z\_118
ump\_119
us\_120
is\_121
g\_122
or\_123
pus\_124
.\_125
Gor\_126
pus\_127
is\_128
earth\_129
y\_130
.\_131
True\_132
or\_133
false\_134
:\_135
Max\_136
is\_137
small\_138
.\_139
Let\_140
us\_141
think\_142
step\_143
by\_144
step\_145
.\_146
\_147
<0x0A>\_148
\#\#\_149
\#\_150
Response\_151
:\_152
<0x0A>\_153
Max\_154
is\_155
zum\_156
pus\_157
.\_158
Z\_159
ump\_160
us\_161
is\_162
g\_163
or\_164
pus\_165
.\_166
Max\_167
is\_168
g\_169
or\_170
pus\_171
.\_172
Gor\_173
p\_174
uses\_175
are\_176
disc\_177
ord\_178
ant\_179
.\_180
Max\_181
is\_182
disc\_183
ord\_184
ant\_185
.\_186
False\_187
<0x0A>\_188
<0x0A>\_189
\#\#\_190
\#\_191
Input\_192
:\_193
<0x0A>\_194
B\_195
or\_196
pin\_197
are\_198
w\_199
ump\_200
us\_201
.\_202
W\_203
ump\_204
uses\_205
are\_206
angry\_207
.\_208
W\_209
ump\_210
us\_211
is\_212
j\_213
empor\_214
.\_215
S\_216
ally\_217
is\_218
le\_219
mp\_220
us\_221
.\_222
L\_223
emp\_224
us\_225
is\_226
w\_227
ump\_228
us\_229
.\_230
True\_231
or\_232
false\_233
:\_234
S\_235
ally\_236
is\_237
fl\_238
oral\_239
.\_240
Let\_241
us\_242
think\_243
step\_244
by\_245
step\_246
.\_247
\_248
<0x0A>\_249
\#\#\_250
\#\_251
Response\_252
:\_253
<0x0A>\_254
S\_255
ally\_256
is\_257
le\_258
mp\_259
us\_260
.\_261
L\_262
emp\_263
us\_264
is\_265
w\_266
ump\_267
us\_268
.\_269
S\_270
ally\_271
is\_272
w\_273
ump\_274
us\_275
.\_276
W\_277
ump\_278
uses\_279
are\_280
angry\_281
.\_282
S\_283
ally\_284
is\_285
angry\_286
.\_287
False\_288
<0x0A>\_289
<0x0A>\_290
\#\#\_291
\#\_292
Input\_293
:\_294
<0x0A>\_295
G\_296
or\_297
pus\_298
is\_299
jel\_300
git\_301
.\_302
Y\_303
ump\_304
uses\_305
are\_306
loud\_307
.\_308
Gor\_309
pus\_310
is\_311
y\_312
ump\_313
us\_314
.\_315
Y\_316
ump\_317
us\_318
is\_319
orange\_320
.\_321
R\_322
ex\_323
is\_324
g\_325
or\_326
pus\_327
.\_328
True\_329
or\_330
false\_331
:\_332
R\_333
ex\_334
is\_335
loud\_336
.\_337
Let\_338
us\_339
think\_340
step\_341
by\_342
step\_343
.\_344
<0x0A>\_345
\#\#\_346
\#\_347
Response\_348
:\_349
<0x0A>\_350
R\_351
ex\_352
is\_353
g\_354
or\_355
pus\_356
.\_357
Gor\_358
pus\_359
is\_360
y\_361
ump\_362
us\_363
.\_364
R\_365
ex\_366
is\_367
y\_368
ump\_369
us\_370
.\_371
Y\_372
ump\_373
uses\_374
are\_375
loud\_376
.\_377
R\_378
ex\_379
is\_380
loud\_381
.\_382
True\_383
<0x0A>\_384
<0x0A>\_385
\#\#\_386
\#\_387
Input\_388
:\_389
<0x0A>\_390
L\_391
emp\_392
us\_393
is\_394
t\_395
ump\_396
us\_397
.\_398
Max\_399
is\_400
le\_401
mp\_402
us\_403
.\_404
T\_405
ump\_406
us\_407
is\_408
f\_409
ru\_410
ity\_411
.\_412
T\_413
ump\_414
us\_415
is\_416
bright\_417
.\_418
L\_419
emp\_420
us\_421
is\_422
drop\_423
ant\_424
.\_425
True\_426
or\_427
false\_428
:\_429
Max\_430
is\_431
bright\_432
.\_433
Let\_434
us\_435
think\_436
step\_437
by\_438
step\_439
.\_440
<0x0A>\_441
\#\#\_442
\#\_443
Response\_444
:\_445
<0x0A>\_446
Max\_447
is\_448
le\_449
mp\_450
us\_451
.\_452
L\_453
emp\_454
us\_455
is\_456
t\_457
ump\_458
us\_459
.\_460
Max\_461
is\_462
t\_463
ump\_464
us\_465
.\_466
T\_467
ump\_468
us\_469
is\_470
bright\_471
.\_472
Max\_473
is\_474
bright\_475
.\_476
True\_477
<0x0A>\_478
<0x0A>\_479
\#\#\_480
\#\_481
Input\_482
:\_483
<0x0A>\_484
S\_485
ter\_486
p\_487
uses\_488
are\_489
d\_490
ull\_491
.\_492
Imp\_493
us\_494
is\_495
medium\_496
.\_497
Imp\_498
uses\_499
are\_500
ster\_501
p\_502
uses\_503
.\_504
W\_505
ren\_506
is\_507
imp\_508
us\_509
.\_510
Ster\_511
pus\_512
is\_513
da\_514
ump\_515
in\_516
.\_517
True\_518
or\_519
false\_520
:\_521
W\_522
ren\_523
is\_524
mel\_525
od\_526
ic\_527
.\_528
Let\_529
us\_530
think\_531
step\_532
by\_533
step\_534
.\_535
<0x0A>\_536
\#\#\_537
\#\_538
Response\_539
:\_540
<0x0A>\_541
W\_542
ren\_543
is\_544
imp\_545
us\_546
.\_547
Imp\_548
uses\_549
are\_550
ster\_551
p\_552
uses\_553
.\_554
W\_555
ren\_556
is\_557
ster\_558
pus\_559
.\_560
Ster\_561
p\_562
uses\_563
are\_564
d\_565
ull\_566
.\_567
W\_568
ren\_569
is\_570
d\_571
ull\_572
.\_573
False\_574
<0x0A>\_575
<0x0A>\_576
\#\#\_577
\#\_578
Input\_579
:\_580
<0x0A>\_581
J\_582
om\_583
pus\_584
is\_585
wooden\_586
.\_587
Imp\_588
us\_589
is\_590
j\_591
om\_592
pus\_593
.\_594
St\_595
ella\_596
is\_597
imp\_598
us\_599
.\_600
True\_601
or\_602
false\_603
:\_604
St\_605
ella\_606
is\_607
wooden\_608
.\_609
Let\_610
us\_611
think\_612
step\_613
by\_614
step\_615
.\_616
<0x0A>\_617
\#\#\_618
\#\_619
Response\_620
:\_621
<0x0A>\_622
St\_623
ella\_624
is\_625
imp\_626
us\_627
.\_628
Imp\_629
us\_630
is\_631
j\_632
om\_633
pus\_634
.\_635
St\_636
ella\_637
is\_638
}


\end{document}